%% file: cvpr.tex
\documentclass[final]{cvpr}

\usepackage{times}
\usepackage{epsfig}
\usepackage{graphicx}
\usepackage{amsmath}
\usepackage{amssymb}

\usepackage[dvipsnames]{xcolor}

\usepackage{graphicx}
\usepackage{comment}
\usepackage{amsmath,amssymb} 
\usepackage{color, colortbl}

\usepackage{xcolor}
\usepackage{nicefrac}
\usepackage{booktabs}
\usepackage{placeins}
\usepackage{pifont}
\usepackage{subcaption}
\usepackage{xspace}
\usepackage{arydshln}
\usepackage{nicefrac}

\usepackage{algorithm}
\usepackage{algpseudocode}
\algrenewcommand\algorithmicindent{1.0em}%

\usepackage{array}
\usepackage{booktabs}
\usepackage{adjustbox}

\usepackage{listings}
\usepackage{multirow}

\usepackage[colorlinks,citecolor=red,urlcolor=blue,bookmarks=false,hypertexnames=true,pagebackref=true,breaklinks=true]{hyperref}

\renewcommand{\backref}[1]{}
\renewcommand{\backrefalt}[4]{%
  \ifcase #1 %
No citations.
  \or
(page #2).%
  \else
(pages #2).%
  \fi%
}

\newcommand{\mypm}{\,$\pm$\,}
\newcommand{\mysmpm}[1]{\scriptsize{\mypm#1}}

\definecolor{BrickRed}{HTML}{B6321C}
\definecolor{RoyalBlue}{HTML}{0071BC}
\definecolor{PineGreen}{HTML}{008B72}
\definecolor{bluefig}{HTML}{5B9BD5}
\definecolor{Gray}{gray}{0.9}

\let\thetaold\theta
\renewcommand{\theta}{\boldsymbol{\thetaold}}

\newcommand{\mcC}{\mathcal{C}}

\newcommand{\pp}{\textit{p.p}\,}

\newcommand{\cmark}{\ding{51}}%
\newcommand{\xmark}{\ding{55}}%

\newcommand\Tstrut{\rule{0pt}{2.6ex}}         
\newcommand\Bstrut{\rule[-0.9ex]{0pt}{0pt}}   

\newcolumntype{R}[2]{%
    >{\adjustbox{angle=#1,lap=\width-(#2)}\bgroup}%
    l%
    <{\egroup}%
}
\newcommand*\rot{\multicolumn{1}{R{45}{1em}}}

\def\cvprPaperID{2527} 
\def\confYear{CVPR 2022}

\begin{document}

\title{DyTox: Transformers for Continual Learning with DYnamic TOken eXpansion}

\author{Arthur Douillard\textsuperscript{1,2}, Alexandre Ramé\textsuperscript{1}, Guillaume Couairon\textsuperscript{1,3}, Matthieu Cord\textsuperscript{1,4}\\
\textsuperscript{1}Sorbonne Université, \textsuperscript{2}Heuritech, \textsuperscript{3}Meta AI, \textsuperscript{4}valeo.ai
\\{\tt\small arthur.douillard@heuritech.com, \{alexandre.rame, matthieu.cord\}@sorbonne-universite.fr,}\\{\tt\small gcouairon@fb.com}
}

\maketitle


\input{content/abstract}
\input{content/intro}
\input{content/related}
\input{content/model}
\input{content/exp}

\input{content/conclusion}

\vspace{0.1cm}{ \noindent\textbf{Acknowledgments}: This work was partly supported by ANR grant VISA DEEP (ANR-20-CHIA-0022), and the HPC  resources  of IDRIS AD011011706.}

\clearpage

\input{content/appendix}

{\small
\bibliographystyle{ieee_fullname}
\bibliography{arthurcitations}
}


\end{document}

%% file: content/abstract.tex
\begin{abstract} 

Deep network architectures struggle to continually learn new tasks without forgetting the previous tasks. A recent trend indicates that dynamic architectures based on an expansion of the parameters can reduce catastrophic forgetting efficiently in continual learning. However, existing approaches often require a task identifier at test-time, need complex tuning to balance the growing number of parameters, and barely share any information across tasks. As a result, they struggle to scale to a large number of tasks without significant overhead.\\
In this paper, we propose a transformer architecture based on a dedicated encoder/decoder framework. Critically, the encoder and decoder are shared among all tasks. Through a dynamic expansion of special tokens, we specialize each forward of our decoder network on a task distribution. Our strategy scales to a large number of tasks while having negligible memory and time overheads due to strict control of the expansion of the parameters. Moreover, this efficient strategy doesn't need any hyperparameter tuning to control the network's expansion. Our model reaches excellent results on CIFAR100 and state-of-the-art performances on the large-scale ImageNet100 and ImageNet1000 while having fewer parameters than concurrent dynamic frameworks.\footnote{\footnotesize{\url{https://github.com/arthurdouillard/dytox}.}}
\end{abstract}

%% file: content/intro.tex
\section{Introduction}

\begin{figure}
    \centering
    \includegraphics[width=0.45\textwidth]{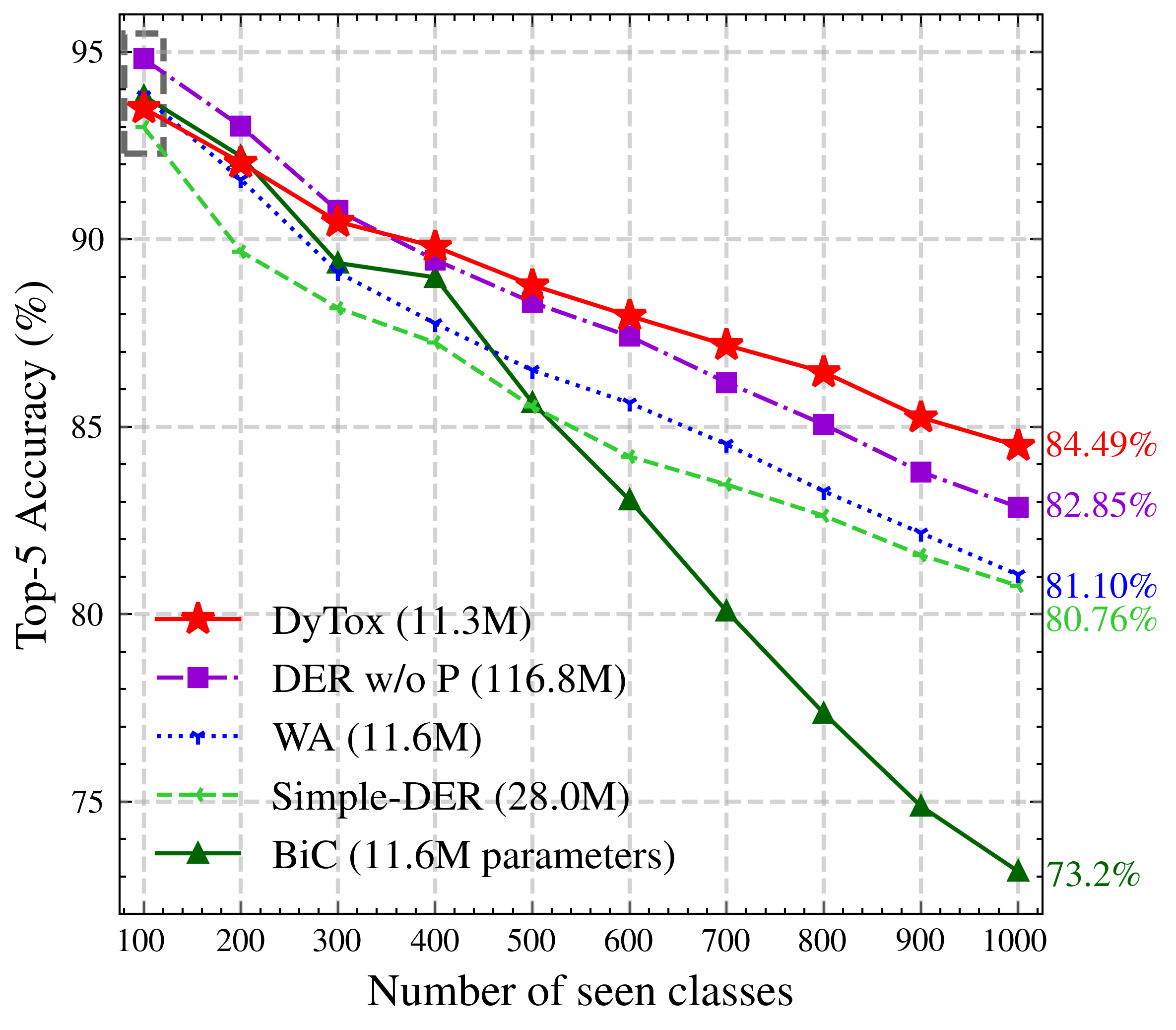}
    \caption{\textbf{DyTox's continual learning performance on ImageNet1000}: for each task, 100 new classes are learned while previously learned classes are not fully accessible but shouldn't be forgotten. Our strategy DyTox (in \textbf{\textcolor{red}{red}}) is state-of-the-art by a large margin. Note that at the initial step before the continual process begins (denoted by a dashed rectangle \includegraphics[width=1.9ex]{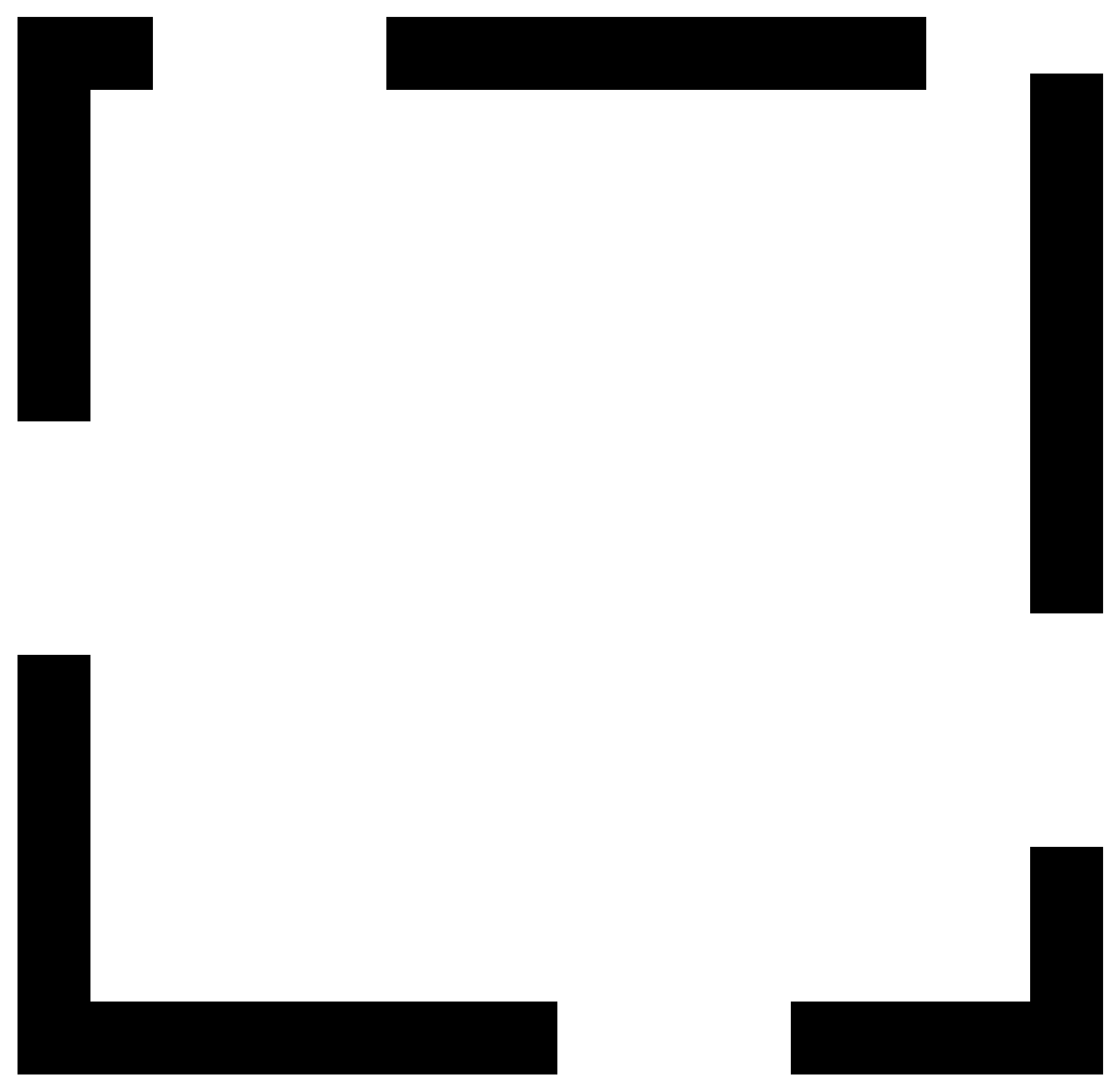}), our model has performance comparable to other baselines: the performance gain is achieved by reducing catastrophic forgetting. Moreover, we have systematically fewer parameters than previous approaches.}
    \label{fig:imagenet1000}
\end{figure}

Most of the deep learning literature focuses on learning a model on a fixed dataset. However, real-world data constantly evolve through time, leading to ever-changing distributions: \textit{i.e.}, new classes or domains appeared. When a model loses access to previous classes data (\textit{e.g.}, for privacy reasons) and is fine-tuned on new classes data, it \textbf{catastrophically forgets} the old distribution. Continual learning models aim at balancing a rigidity/plasticity trade-off where old data are not forgotten (rigidity to changes) while learning new incoming data (plasticity to adapt). Despite recent advances, it is still an open challenge.

A growing amount of efforts have emerged to tackle catastrophic forgetting \cite{rebuffi2017icarl,kirkpatrick2017ewc,wu2019bias_correction,hou2019ucir,douillard2020podnet,yan2021der}. Recent works \cite{yoon2018dynamically_expandable_networks,li2019learning_to_grow,hung2019cpg,fernando2017path_net,golkar2019neural_pruning,serra2018hat} dynamically expand the network architectures \cite{yoon2018dynamically_expandable_networks,li2019learning_to_grow} or re-arrange their internal structures \cite{fernando2017path_net,serra2018hat,hung2019cpg,golkar2019neural_pruning}. Unfortunately at test-time, they require to know the task to which the test sample belongs --- in order to know which parameters should be used. More recently, DER \cite{yan2021der} and Simple-DER \cite{li2021preserve} discarded the need for this task identifier by learning a single classifier on the concatenation of all produced embeddings by different subsets of parameters. Yet, these strategies induce dramatic memory overhead when tackling a large number of tasks, and thus need complex pruning as post-processing.

To improve the ease of use of continual learning frameworks for real-world applications, we aim to design a dynamically expandable representation (almost) `for free' by having the three following properties: \#1 \textbf{limited memory overhead} as the number of tasks grows, \#2 \textbf{limited time overhead} at test time and \#3 \textbf{no setting-specific hyperparameters} for improved robustness when faced to an unknown (potentially large) number of tasks.

To this end, we leverage the computer vision transformer ViT \cite{dosovitskiy2020vit}. Transformers \cite{vaswani2017transformer}  offer a very interesting framework to satisfy the previously mentioned constraints. Indeed, we build upon this architecture to design a \textbf{encoder/decoder strategy}: the encoder layers are shared among all members of our dynamic network; the unique decoder layer is also shared but its forward pass is specialized by a \textbf{task-specific learned token} to produce task-specific embeddings. Thus, the memory growth of the dynamic network is extremely limited: only a 384d vector per task, validating property \#1. Moreover, this requires no hyperparameter tuning (property \#3). Finally, the decoder is explicitly designed to be computationally lightweight (satisfying property \#2).
We nicknamed our framework, DyTox, for \textbf{DYnamic TOken eXpansion}. To the best of our knowledge, we are the first to apply the transformer architecture to continual computer vision. 

Our strategy is robust to different settings, and can easily scale to a large number of tasks. In particular, we validate the efficiency of our approach on CIFAR100, ImageNet100, and ImageNet1000 (displayed on \autoref{fig:imagenet1000}) for multiple settings. We reach state-of-the-art results, with only a small overhead thanks to our efficient dynamic strategy.

%% file: content/related.tex
\section{Related work}

\paragraph{Continual learning} models tackle the catastrophic forgetting of the old classes \cite{thrun1998lifelonglearning,french1999catastrophicforgetting}. In computer vision, most of continual learning strategies applied on large-scale datasets use rehearsal learning: a limited amount of the training data of old classes is kept during training \cite{robins1995catastrophicforgetting}. This data is usually kept in raw form (\textit{e.g.}, pixels) \cite{rebuffi2017icarl,castro2018end_to_end_inc_learn,chaudhry2019tinyepisodicmemories} but can also be compressed \cite{hayes2020remind,iscen2020incrementalfeatureadaptation}, or trimmed \cite{douillard2021objectrehearsal} to reduce memory overhead; others store only a model to generate new samples of past classes \cite{kemker2018fearnet,shin2017deep_generative_replay,lesort2019generative}.
In addition, most approaches aim at limiting the changes in the model when new classes are learned. These constraints can be directly applied on the weights \cite{kirkpatrick2017ewc,zenke2017synaptic_intelligence,aljundi2018MemoryAwareSynapses,chaudhry2018riemannien_walk}, intermediary features \cite{hou2019ucir,dhar2019learning_without_memorizing_gradcam,peng2019m2kd,douillard2020podnet,douillard2020plop}, prediction probabilities \cite{li2018lwf,rebuffi2017icarl,castro2018end_to_end_inc_learn,cermelli2020modelingthebackground}, or on the gradients \cite{lopezpaz2017gem,chaudhry2019AGEM,farajtabar2020ogd,saha2021gpm}. All these constraint-based methods use the same static network architectures which doesn't evolve through time, usually a ResNet \cite{he2016resnet}, a LeNet \cite{lecun1999lenet}, or a small MLP.

\paragraph{Continual dynamic networks} In contrast, our paper and others focus on designing \textbf{dynamic architectures} that best handle a growing training distribution \cite{yoon2018dynamically_expandable_networks,li2019learning_to_grow}, in particular by dynamically creating (sub-)members each specialized in one specific task \cite{fernando2017path_net,golkar2019neural_pruning, hung2019cpg,rusu2016progressive,routingnetworkcollier,wen2020batchensemble}. Unfortunately, previous approaches often require the sample's task identifier at test-time to select the right subset of parameters. We argue this is an unrealistic assumption in a real-life situation where new samples could come from any task.
Recently, DER \cite{yan2021der} proposed
a dynamic expansion of the representation by adding a new feature extractor per task. All extractors' embeddings would then be concatenated and fed to a unified classifier, discarding the need for a task identifier at test-time. To limit an explosion in the number of parameters, they aggressively prune each model after each task using the HAT \cite{serra2018hat} procedure. Unfortunately, the pruning is hyperparameter sensitive. Therefore, hyperparameters are tuned differently on each experiment: for example, learning a dataset in 10 steps or in 50 steps use different hyperparameters. While being impracticable, it is also unrealistic because the number of classes is not known in advance in a true continual situation. Simple-DER \cite{li2021preserve} also uses multiple extractors, but its pruning method doesn't need any hyperparameters; the negative counterpart is that Simple-DER controls less the parameter growth (2.5x higher than a base model).
In contrast, we propose a framework dedicated to continual learning that seamlessly enables a task-dynamic strategy, efficient on all settings, without any setting-dependant modification and at almost no memory overhead. We share early class-agnostic \cite{olah2017feature} layers similarly to TreeNets \cite{lee2015treenet} and base our strategy on the Transformer architecture.

\vspace{-1em}

\paragraph{Transformers} were first introduced for machine translation \cite{vaswani2017transformer}, with the now famous self-attention.
While the original transformer was made of encoder and decoder layers, later transformers starting from BERT \cite{devlin2018bert} used a succession of identical encoder blocks. Then, ViT \cite{dosovitskiy2020vit} proposed to apply transformers to computer vision by using patches of pixels as tokens. Multiple recent works, including DeiT \cite{touvron2021deit}, CaiT \cite{touvron2021cait}, ConVit \cite{dascoli2021convit}, and Swin \cite{liu2021swin}, improved ViT with architecture and training procedures modifications. PerceiverIO \cite{jaegle2021perceiverio} proposed a general architecture whose output is adapted to different modalities using specific learned tokens, and whose computation is reduced using a small number of latent tokens. Despite being successful across various benchmarks, transformers have not yet been considered for continual computer vision to the best of our knowledge.
Yet, we don't use the transformer architecture for its own sake, but rather because of the intrinsic properties of transformers; in particular, the seminal encoder/decoder framework allows us to build an efficient architecture with strong capabilities against catastrophic forgetting.

%% file: content/model.tex
\begin{figure*}
    \centering
    \includegraphics[width=0.95\textwidth]{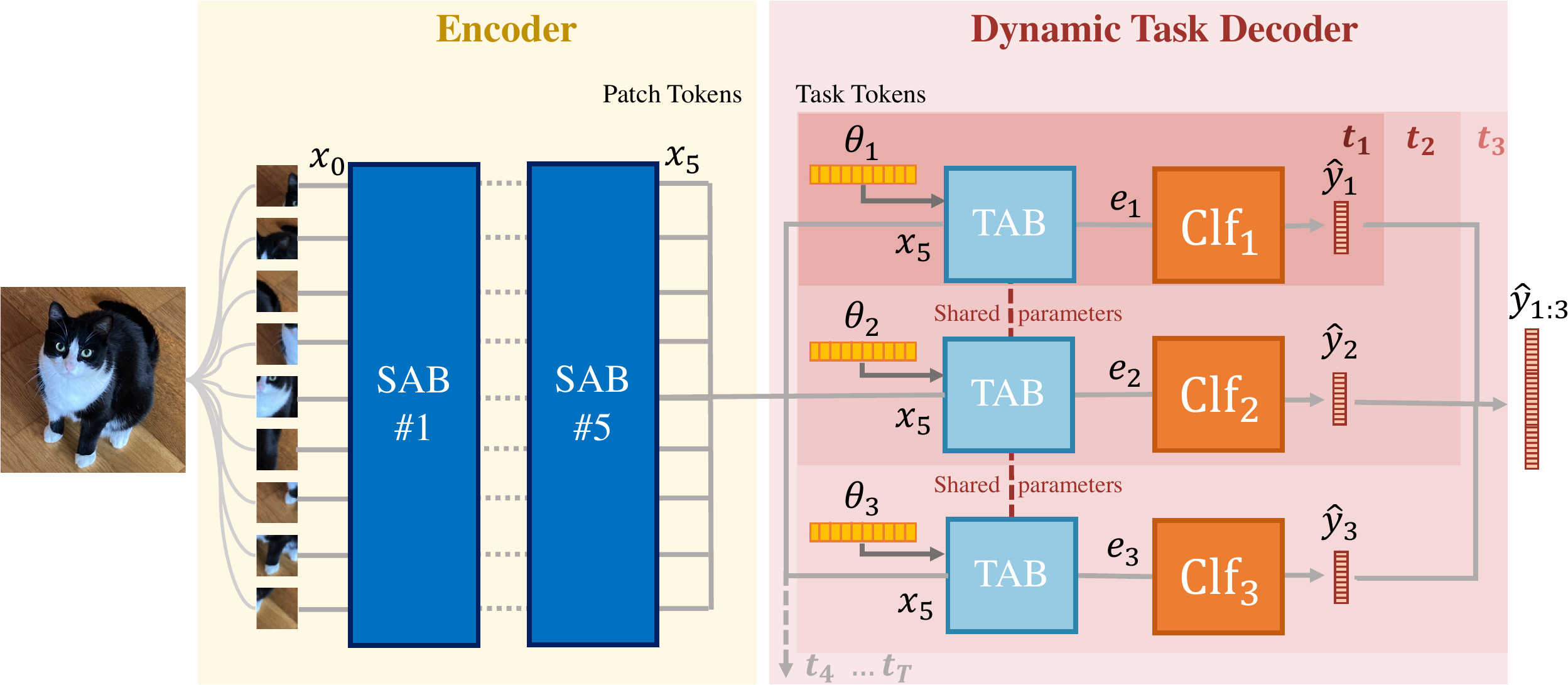}
    \caption{\textbf{DyTox transformer model}. An image is first split into multiple patches, embedded with a linear projection. The resulting patch tokens are processed by 5 successive Self-Attention Blocks (SAB) (\autoref{sec:vit}). For each task ($t = 1\dots T$), the processed patch tokens are then given to the Task-Attention Block (TAB) (\autoref{sec:tab}): each forward through the TAB is modified by a different task-specialized token $\theta_t\, \text{for}\, t \in \{1 \dots T\}$ (\autoref{sec:ensembling_cab}). The $T$ final embeddings are finally given separately to independent classifiers $\text{Clf}_t$ each predicting their task's classes $C^t$. All $|C^{1:T}|$ logits are activated with a sigmoid. For example, at task $t=3$, one forward is done through the SABs and three task-specific forwards through the unique TAB.}
    \label{fig:tense_model}
\end{figure*}

\section{DyTox transformer model}
\label{sec:model}

\label{sec:problem}

Our goal is to learn a unified model that will classify an increasingly growing number of classes, introduced in a fixed amount of steps $T$.
At a given step $t \in \{1 \dots T\}$, the model is exposed to new data belonging to new classes. Specifically, it learns from samples $\{(x_i^t, y_i^t)\}_{i}$, where $x_i^t$ is the $i$-th image of this task $t$ and $y_i^t$ is the associated label within the label set $\mcC^t$. All task label sets are exclusive: $\mcC^0 \cap \mcC^1 \dots \mcC^T = \emptyset$.
The main challenge is that the data are fully available only temporarily: following most previous works, only a few samples from previous tasks $\{1 \dots t-1\}$ are available for training at step $t$ as rehearsing data. Yet, the model should remain able to classify test data coming from all seen classes $\mcC^{1:t}$. A table of notations is provided in the supplementary materials.

The \autoref{fig:tense_model} displays our DyTox framework, which is made of several components (SAB, TAB, and Task Tokens) that we describe in the following sections. 

\subsection{Background}
\label{sec:vit}

The vision transformer \cite{dosovitskiy2020vit} has three main components: the patch tokenizer, the encoder made of Self-Attention Blocks, and the classifier.

\begin{figure}
    \centering
    \includegraphics[width=0.45\textwidth]{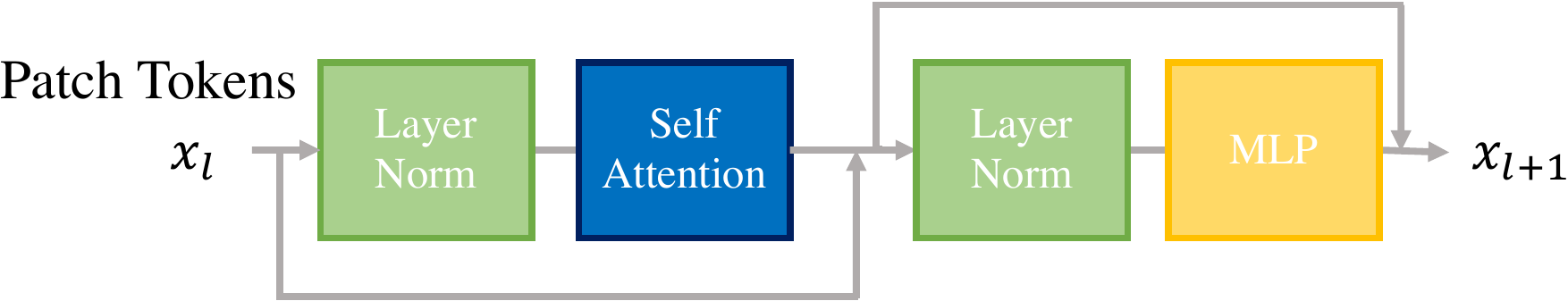}
    \caption{\textbf{The Self-Attention Block (SAB)} combines a Self-Attention (SA), two Layer Norms, and one MLP with a single hidden layer. As in a ResNet, two shortcuts are used with element-wise addition.\vspace{-2em}}
    \label{fig:sab}
\end{figure}

\paragraph{Patch tokenizer}The fixed-size input RGB image is cropped into $N$ patches of equal dimensions and then projected with a linear layer to a dimension $D$. Both operations, the cropping and projection, are done with a single 2D convolution whose kernel size is equal to its stride size. The resulting tensor $x_0 \in \mathbb{R}^{N \times D}$ is extended with a learned class token $x_{\text{cls}} \in \mathbb{R}^D$ resulting in a tensor of shape $\mathbb{R}^{(N+1) \times D}$. Following \cite{gehring2017posembeddings}, a learned positional embedding $p \in \mathbb{R}^{(N+1) \times D}$ is added (element-wise).

\paragraph{Self-Attention (SA) based encoder}The tokens are fed to a stack of transformer blocks that we denote here as Self-Attention Blocks (SABs):
\begin{equation}
\begin{aligned}
x_{l}^{\prime} &=x_{l}+\operatorname{SA}_l\left(\operatorname{Norm}_{l,1}\left(x_{l}\right)\right)\,, \\
x_{l+1} &=x_{l}^{\prime}+\operatorname{MLP}_l\left(\operatorname{Norm}_{l,2}\left(x_{l}^{\prime}\right)\right),
\end{aligned}
\label{eq:sa_block}
\end{equation}
with $\operatorname{SA}$ a Self-Attention layer \cite{vaswani2017transformer}, $\operatorname{Norm}$ a layer normalization \cite{ba2016layernorm}, and $\operatorname{MLP}$ a Multi-Layer Perceptron with a single hidden layer. We repeat these operations for each SAB, from $l=1$ to $l=L$. The resulting tensor (which keeps the same dimension after every block) is $x_{L} \in \mathbb{R}^{(N+1) \times D}$. We display a visual illustration of a SA Block in \autoref{fig:sab}.

\paragraph{Classifier} In the original vision transformer (ViT \cite{dosovitskiy2020vit}), a learned vector called the ``\textit{class token}'' is appended to the patch tokens after the tokenizer. This special class token, when processed after all the SABs, is given to a linear classifier with a softmax activation to predict the final probabilities. However, more recent works, as CaiT \cite{touvron2021cait}, propose instead to introduce the class token only at the ultimate or penultimate SAB to improve classification performance.

\subsection{Task-Attention Block (TAB)}
\label{sec:tab}

Contrary to previous transformer architectures, we don't have a class token, but rather what we nicknamed ``\textbf{task tokens}''; the learned token of the $i^{th}$ task is denoted $\theta_i$. This special token will only be added at the last block. To exploit this task token, we define a new attention layer, that we call the Task-Attention. It first concatenates the patch tokens $x_L$ produced by the ultimate SAB with a task token $\theta_i$:
\begin{equation}
z_i = [\theta_i, x_L] \, \in \mathbb{R}^{(\mathbb{N} + 1) \times \mathbb{D}}\,.
\label{eq:concat_cls_token}
\end{equation}
This is then given to the Task-Attention (TA), inspired by the Class-Attention of Touvron et al. \cite{touvron2021cait}:
\begin{equation}
\begin{aligned}
Q_i &=W_{q} \theta_i\,, \\
K_i &=W_{k} z_i\,, \\
V_i &=W_{v} z_i\,, \\
A_i &=\operatorname{Softmax}\left(Q_i \cdot K_i^{T} / \sqrt{d / h}\right)\,,\\
O_i &= W_{o} A_i V_i+b_{o} \, \in \mathbb{R}^{1 \times \mathbb{D}}\,,
\end{aligned}
\label{eq:ca_layer}
\end{equation}
with $d$ being the embedding dimension, and $h$ the number of attention heads \cite{vaswani2017transformer}. Contrary to the classical Self-Attention, the Task-Attention defines its query ($Q_i$) only from the task-token $\theta_i$ without using the patch tokens $x_L$. The Task-Attention Block (TAB) is then a variation of the SAB where the attention is a Task-Attention (TA):
\begin{equation}
\begin{aligned}
c^{\prime} &=c+\operatorname{TA}\left(\operatorname{Norm}_1\left(z\right)\right)\,, \\
c^{\prime\prime} &=c^{\prime}+\operatorname{MLP}\left(\operatorname{Norm}_2\left(c^{\prime}\right)\right)\,.
\end{aligned}
\label{eq:ca_block}
\end{equation}
Overall, our new architecture can be summarized by the repetition of SA Blocks $\{\operatorname{SAB}_l\}_{l=1}^{L}$ (defined in \autoref{eq:sa_block}) ended by a single TA Block $\operatorname{TAB}$ (defined in \autoref{eq:ca_block}):
\begin{equation}
    e_i = \operatorname{TAB} \circ\, ([\theta_i,\, \operatorname{SAB}_{l=L} \circ\, ... \operatorname{SAB}_{l=1}(x_0)]) \in \mathbb{R}^D\,.
\label{eq:cab_sab}
\end{equation}
The final embedding $e_i$ is fed to a classifier $\operatorname{clf}$ made of a $\operatorname{Norm}_c$ and a linear projection parametrized by $\{W_c, b_c\}$:
\begin{equation}
    \tilde{y}_i = \operatorname{Clf}(e_i) = W_c \operatorname{Norm}_c(e_i) + b_c\,.
\end{equation}

\subsection{Dynamic task token expansion}
\label{sec:ensembling_cab}
We defined in the previous section our base network, made of a succession of SABs and ended by a single TAB. As detailed, the TAB has two inputs: the patch tokens $x_L$ extracted from the image and a learned task-token $\theta_i$. We'll now detail how our framework evolves in a continual situation at each new step.

During the first step, there is only one task token $\theta_1$. At each new step, we propose to expand our parameter space by creating a new task token while keeping the previous ones. Thus, after $t$ steps, we have $t$ task tokens ($\theta_i\,\text{for}\, i\in \{1 \dots t\}$). Given an image $x$ --- belonging to any of the seen tasks $\{1\dots \, t\}$ --- our model tokenizes it into $x_0$, and processes it through the multiple SABs: this outputs the patch tokens $x_L$. Finally, our framework does as many forward passes through the TAB as there are tasks: critically, each TAB forward passes is executed with a different task token $\theta_i$, resulting in different task-specific forwards, each producing the task-specific embeddings $e_i$ (see \autoref{fig:tense_model}):
\begin{equation}
\begin{aligned}
    & e_1 = \operatorname{TAB}([\theta_1, x_L])\,,\\
    & e_2 = \operatorname{TAB}([\theta_2, x_L])\,,\\
    & \dots\\
    & e_t = \operatorname{TAB}([\theta_t, x_L])\,.\\
\end{aligned}
\label{eq:multiple_tab}
\end{equation}
Rather than concatenating all embeddings $\{e_1, e_2, \dots, e_t\}$ together and feeding them to one classifier, we leverage \textbf{task-specific classifiers}. Each classifier $\operatorname{clf}_i$ is made of a $\operatorname{Norm}_i$ and a linear projection parametrized by $\{W_i, b_i\}$, with $W_i \in \mathbb{R}^{\mcC^i \times D}$ and $b \in \mathbb{R}^{\mcC^i}$. It takes as input its task-specific embedding $e_i$ and returns:
\begin{equation}
    \hat{y}_i = \operatorname{Clf}_i(e_i) = \sigma(W_i \operatorname{Norm}_i e_i + b_i)\,,
\end{equation}
the predictions for the classes $y_i \in \mcC^i$, where $\sigma(x) = \nicefrac{1}{(1 + e^{-x})}$ is the sigmoid activation. In comparison with the softmax activation, the element-wise sigmoid activation reduces the overconfidence in recent classes. Consequently, the model is better calibrated, which is an important attribute of continual model \cite{belouadah2019il2m,wu2019bias_correction,zhao2020weightalignement}. The loss is the binary-cross entropy. The independent classifiers paradigm \cite{shanahan2021ensembleencoders} coupled with the sigmoid activation and binary cross-entropy loss exclude explicitly a late fusion \cite{ramachandram2017multimodalreview} of the task embeddings resulting in more \textbf{specialized classifiers}.

\paragraph{The overall structure of the DyTox strategy} is illustrated in \autoref{fig:tense_model}. We also show in \autoref{algo:cab_ensemble_1-1_bce} the pseudo-code of a forward pass at test-time after having learned the task $t$. Critically, the test image can belong to any of the previously seen tasks $\{1 \,\dots \, t\}$.
Our dynamic task token expansion is more efficient than a naive parameter expansion that would create a new copy of the whole network for each new task. (1) Our expansion is limited to a new task token per new task, which is only $d=384$ new parameters. This is small compared to the total model size ($\approx$ 11 million parameters). The \textbf{memory overhead is thus almost null}. (2) The computationally intensive blocks (\textit{i.e.}, the SABs) are executed only once despite learning multiple tasks. In contrast, the TAB has as many forwards as there are tasks. Though, this induces minimal overhead because the \textbf{Task-Attention has a linear complexity w.r.t the number of patches} while the Self-Attention is quadratic. Therefore, the time overhead is sub-linear. We quantitatively show this in \autoref{sec:exp}.

\begin{algorithm}[tb]
\caption{DyTox's forward pass at step $t$}
\label{algo:cab_ensemble_1-1_bce}
\hspace*{\algorithmicindent} \textbf{Input:} $x_0$ (initial patch tokens), $y$ ( ground-truth labels) \\
 \hspace*{\algorithmicindent} \textbf{Output:} $\hat{y}_{1:t}$ (predictions for all classes of $\mcC^{1:t}$)
\begin{algorithmic}[1]
 \State $x_L \gets \operatorname{SAB}_{l=L} \circ ... \operatorname{SAB}_{l=1}(x_0)$  \Comment{\autoref{sec:vit}} 
 
 \For{\texttt{$i \gets 1$; $i \leq t$; $i{+}{+}$}}
    \State $e_i \gets \operatorname{TAB}([\theta_i, x_L])$ \Comment{\autoref{sec:tab}}
    \State $\hat{y}_i \gets \operatorname{Clf}_i(e_i)$  \Comment{\autoref{sec:ensembling_cab}}
 \EndFor 
 
 \State $\hat{y}_{1:t} \gets [\hat{y}_1,\, \dots,\, \hat{y}_{t}]$
\end{algorithmic}
\end{algorithm}

\vspace{-1em}
\paragraph{Context} The current transformer paradigm starting from BERT \cite{devlin2018bert} and continuing with ViT \cite{dosovitskiy2020vit} is based on a encoder+classifier structure. Differently, our dynamic framework strays is a resurgence of the encoder/decoder structure of the original transformer \cite{vaswani2017transformer}: the encoder is shared (both in memory and execution) for all outputs. The decoder parameters are also shared, but its execution is task-specific with each task token, with each forward akin to a task-specific expert chosen from a mixture of experts \cite{masoudnia2014mixture}. Moreover, multi-tasks text-based transformers have natural language tokens as an indicator of a task \cite{raffel2019t5} (\textit{e.g.} "summarize the following text"), in our context of vision we used our defined task tokens as indicators.

\label{sec:training}

\vspace{-0.5em}
\paragraph{Losses} Our model is trained with three losses: (1) the classification loss $\mathcal{L_\text{clf}}$, a binary-cross entropy, (2) a knowledge distillation \cite{hinton2015knowledge_distillation} $\mathcal{L_\text{kd}}$ applied on the probabilities, and (3) the divergence loss $\mathcal{L_\text{div}}$. The distillation loss helps to reduce forgetting. It is arguably quite naive, and more complex distillation losses \cite{selvaraju2017gradcam,hou2019ucir,douillard2020podnet} could further improve results. The divergence loss, inspired from the ``auxiliary classifier'' of DER \cite{yan2021der}, uses the current last task's embedding $e_t$ to predict ($|\mcC^t| + 1$) probabilities: the current last task's classes $\mcC^t$ and an extra class representing all previous classes that can be encountered via rehearsal. This classifier is discarded at test-time and encourages a better diversity among task tokens. The total loss is:
\begin{equation}
    \mathcal{L} = (1 - \alpha) \mathcal{L_\text{clf}} + \alpha \mathcal{L_\text{kd}} + \lambda \mathcal{L_\text{div}}\,,
\label{eq:final_loss}
\end{equation}
with $\lambda$ a hyperparameter set to $0.1$ for \textbf{all} experiments. $\alpha$ correspond to the fraction of the number of old classes over the number of new classes $\frac{|C^{1:t-1}|}{|C^{1:t}|}$ as done by \cite{zhao2020weightalignement}. Therefore, $\alpha$ is automatically set; this removes the need to finely tune this hyperparameter.

%% file: content/exp.tex
\section{Experiments}
\label{sec:exp}

\subsection{Benchmarks \& implementation}

\paragraph{Benchmarks \& Metrics} We evaluate our model on CIFAR100 \cite{krizhevskycifar100}, ImageNet100 and ImageNet1000 \cite{deng2009imagenet} (descriptions in the supplementary materials) under different settings.
The standard continual scenario in ImageNet has 10 steps: thus we add 10 new classes per step in ImageNet100, and 100 new classes per step in ImageNet1000.
In CIFAR100, we compare performances on 10 steps (10 new classes per step), 20 steps (5 new classes per step), and 50 steps (2 new classes per step).
In addition to the top-1 accuracy, we also compare the top-5 accuracy on ImageNet.
We report the ``\textit{Avg}'' accuracy which is the average of the accuracies after each step as defined by \cite{rebuffi2017icarl}.
We also report the final accuracy after the last step (``\textit{Last}''). Finally, in our tables, ``\textit{\#P}'' denotes the parameters count in million after the final step.

\input{tables/transf_archi}

\input{tables/imagenet}

\paragraph{Implementation details} As highlighted in \autoref{tab:archi}, our network has the same structure across all tasks.
Specifically, we use 5 Self-Attention Blocks (SABs), 1 Task-Attention Block (TAB). All 6 have an embedding dimension of 384 and 12 attention heads.
We designed this shallow transformer to have a comparable parameters count to other baselines, but also made it wider than usual "tiny" models \cite{dosovitskiy2020vit,touvron2021deit,touvron2021cait}.
We tuned all hyperparameters for CIFAR100 with 10 steps on a validation set made of 10\% of the training set, and then kept them fixed for all other settings, ImageNet included.
The only difference between the two datasets is that ImageNet images are larger; thus the patch size is larger, and overall the base transformer has slightly more parameters on ImageNet than on CIFAR (11.00M vs 10.72M) because of a bigger positional embedding.
We use the attention with spatial prior (introduced by ConViT \cite{dascoli2021convit}) for all SABs which allows training transformers on a small dataset (like CIFAR) without pretraining on large datasets or complex regularizations.
Following previous works \cite{rebuffi2017icarl,yan2021der}, we use for all models (baselines included) 2,000 images of rehearsal memory for CIFAR100 and ImageNet100, and 20,000 images for ImageNet1000.
The implementations of the continual scenarios are provided by Continuum \cite{douillardlesort2021continuum}. Our network implementation is based on the DeiT \cite{touvron2021deit} code base which itself uses extensively the timm library \cite{wightman2019timm}. The code is released publicly\footnote{\footnotesize{\url{https://github.com/arthurdouillard/dytox}.}}. The full implementation details are in the appendix.

\begin{figure}
    \centering
    \includegraphics[width=0.40\textwidth]{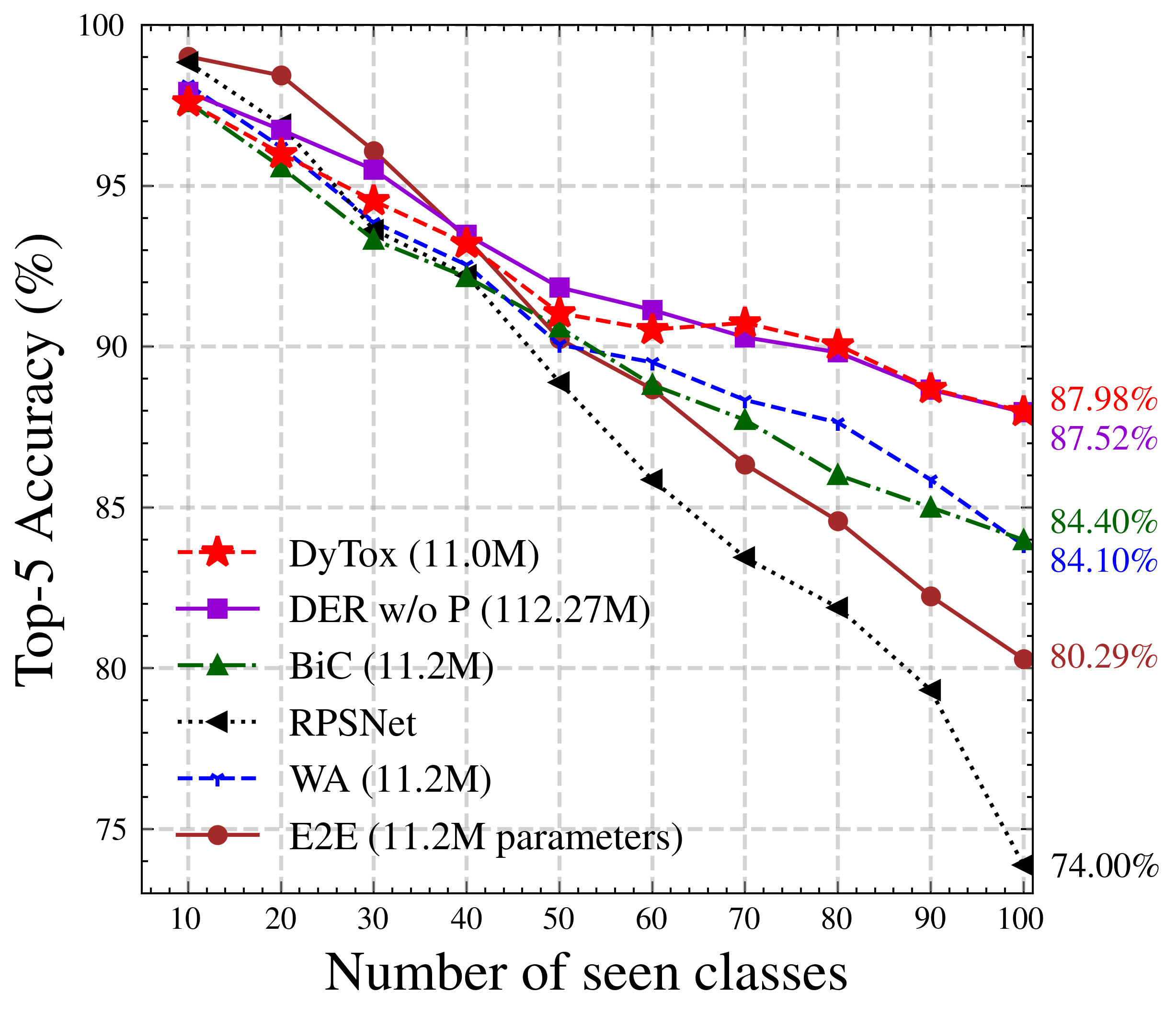}
    \caption{\textbf{Performance evolution on ImageNet100}. The top-5 accuracy (\%) is reported after learning each task. Our model DyTox (in \textbf{\textcolor{red}{red}}) surpasses significantly most baselines, and is of equal performance as the complex DER that uses pruning with setting-specific hyperparameters.\vspace{-1em}}
    \label{fig:imagenet100}
\end{figure}

\subsection{Quantitative results}

\paragraph{ImageNet}
We report performances in \autoref{tab:imagenet} on the complex ImageNet dataset. The $^\dagger$ marks the DER with setting-specific pruning, and DER w/o P is for the DER without pruning. In ImageNet100, DyTox reaches 69.10\% and outperforms DER$^\dagger$ by +3.04 percentage points (\pp) in ``Last'' top-1 accuracy. Though, DyTox and DER w/o P somehow perform similarly in ``Avg'' accuracy on this setup, as highlighted in the performance evolution displayed in \autoref{fig:imagenet100}.
Most importantly, on the larger-scale ImageNet1000, DyTox systematically performs best on all metrics despite having lower parameters count.
Specifically, DyTox reaches 71.29\% in ``Avg'' top-1 accuracy, and 63.34\% in ``Last'' top-1 accuracy. This outperforms the previous state-of-the-art DER w/o P (68.84\% in ``Avg'', 60.16\% in ``Last'') which has 10 ResNet18 in parallel and 116.89M parameters.
Compared to the pruned DER$^\dagger$, DyTox has a +4.56 \pp in top-1 and a +1.51 \pp in top-5 for the ``Avg'' accuracy.
All models evolutions on ImageNet1000 are illustrated in \autoref{fig:imagenet1000}: DyTox constantly surpasses previous state-of-the-art models --- despite having a comparable performance at the first step and fewer parameters.

\input{tables/cifar100_b0}

\begin{figure*}[t!]
\centering
  \includegraphics[width=0.9\linewidth]{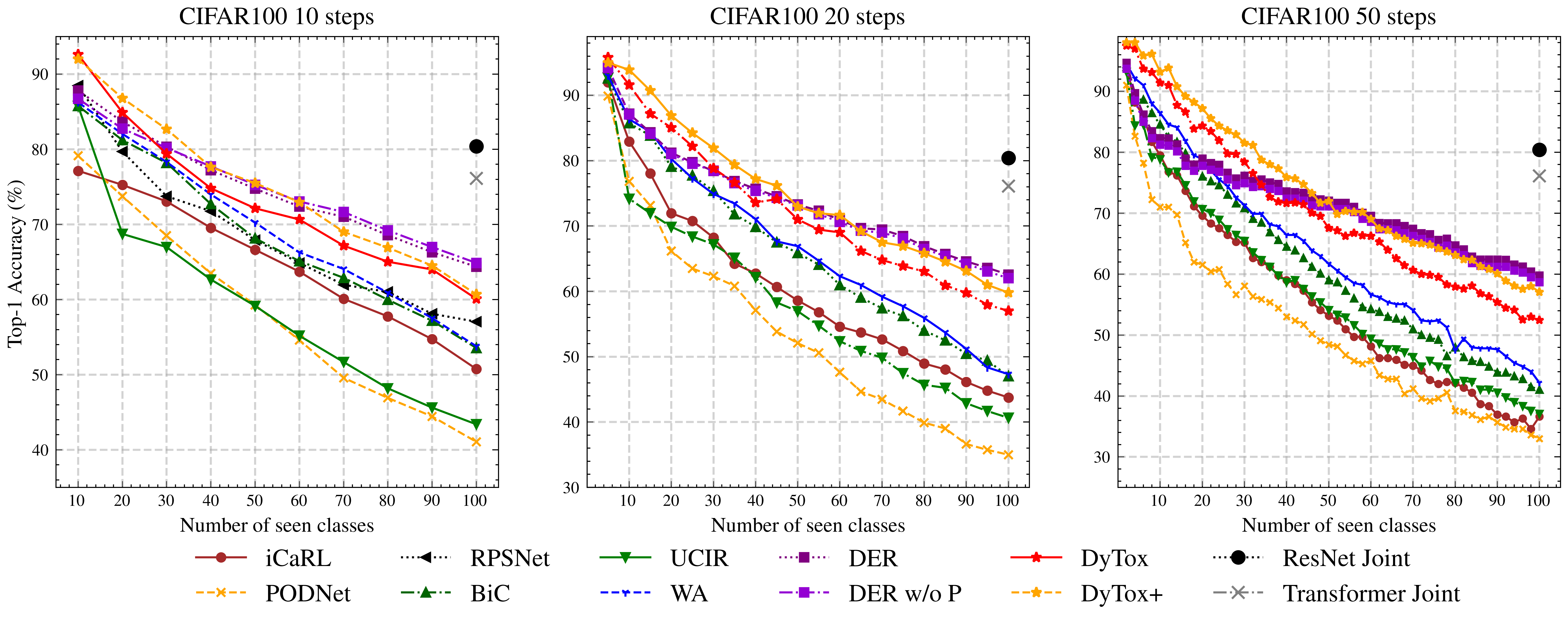}
    \caption{\textbf{Performance evolution on CIFAR100}. The top-1 accuracy (\%) is reported after learning each task. \textbf{Left} is evaluated with 10 steps, \textbf{middle} with 20 steps, and \textbf{right} with 50 steps.}
\label{fig:increment_cifar}
\end{figure*}

DyTox is able to scale correctly while handling seamlessly the parameter growth by sharing most of the weights across tasks.
In contrast, DER had to propose a complex pruning method; unfortunately, this pruning required different hyperparameter values for different settings.
Despite this, the pruning in DER$^\dagger$ is less efficient when classes diversity increase: DER$^\dagger$ doubles in size between ImageNet100 and ImageNet1000 (\cite{yan2021der} reports 7.67M \textit{vs.} 14.52M) while handling the same amount of tasks (10).
Note that these parameter counts reported for DER$^\dagger$ in \cite{yan2021der} are in fact averages over all steps: the final parameters count (necessarily higher) was not available and thus is not reported in our tables. Simple-DER also applies pruning but without hyperparameter tuning; while simpler, the pruning is also less efficient and induces larger model (28.00M parameters). 

\vspace{-0.5em}
\paragraph{CIFAR100} \autoref{tab:cifar100-b0} shows results for all approaches on CIFAR100.
The more steps there are, the larger the forgetting is and thus the lower the performances are.
Those settings are also displayed in \autoref{fig:increment_cifar} after each task.
In every setting, DyTox is close to DER w/o P  for much fewer parameters (up to 52x less). Critically, DyTox is significantly above other baselines: \textit{e.g.} DyTox is up to +25\% in ``Last'' accuracy in the 50 steps setup.

\vspace{-1em}
\paragraph{Improved training procedure} To bridge the gap between DyTox and DER w/o P on CIFAR100, we introduce a new efficient training procedure for continual learning.
Using MixUp \cite{hingyi2018mixup}, we linear interpolate new samples with existing samples. The interpolation factor $\lambda \sim \operatorname{Beta}(\alpha, \alpha)$ is sampled with $\alpha=0.8$: the pixels of two images are mixed ($x = \lambda x_1 + (1 - \lambda) x_2$) as their labels ($y = \lambda y_1 + (1 - \lambda) y_2$). MixUp was shown to have two main effects: (1) it diversifies the training images and thus enlarges the training distribution on the vicinity of each training sample \cite{chapelle2001vicinalrisk} and (2) it improves the network calibration \cite{guo2017miscalibration,thulasidasan2019mixupcalibration}, reducing the overconfidence in recent classes.
Thus MixUp has shared motivation with the sigmoid activation.
When DyTox is combined with this MixUp procedure, nicknamed as DyTox+, this significantly improves the state-of-the-art in ``Avg'' accuracy in all three settings of \autoref{tab:cifar100-b0}. We also provide in the appendix further improvement for this new continual training procedure providing even larger gain on both CIFAR100 and ImageNet100.

\subsection{Model introspection on CIFAR100}

\paragraph{Memory overhead}
We only add a vector of size $d=384$ per task; thus, the overhead in memory (not considering the growing classifier which is common for all continual models) is only of $+0.004\%$ per step. Even in the challenging setting of CIFAR100 with 50 tasks, our memory overhead is almost null ($+0.2\%$).

\label{sec:comp_over}
\vspace{-1em}
\paragraph{Computational overhead} The vast majority of the computation is done in the SABs, thus shared among all tasks. The dynamical component of our model is located at the ultimate TAB. Moreover, the Task-Attention, contrary to the Self-Attention, has a time complexity linear in terms of tokens and not quadratic reducing the time overhead to an acceptable sub-linear amount. Overall, for each new task, one forward pass takes $2.24\%$ more time than for the base transformer.

\vspace{-1em}
\input{tables/plus}

\paragraph{Training procedure introspection} Our DyTox+ strategy with MixUp really reduces catastrophic forgetting and does not just improve raw performances.
This is shown in \autoref{tab:training_plus}, where we compare DyTox \textit{vs.} DyTox+ strategies on CIFAR100.
While MixUp only slightly improves by 1.39 \pp the accuracy in joint learning (no continual, 1 step), MixUp greatly improves the performance by 4.75 \pp in the 50 steps continual scenario.
To further illustrate this, we also report the Chaudhry et al.'s forgetting \cite{chaudhry2018riemannien_walk} measure which compares how performances dropped compared to previous steps. MixUp reduces this forgetting by 1.65 \pp.

\vspace{-1em}
\label{sec:ablations}
\paragraph{Model ablations} We ablate the importance of the different components of DyTox in \autoref{tab:ablation}.
We add on the base transformer a naive knowledge distillation \cite{hinton2015knowledge_distillation} and a finetuning \cite{castro2018end_to_end_inc_learn,hou2019ucir,douillard2020podnet,yan2021der} applied after each task on a balanced set of new data and rehearsal data.
Finally, our DyTox strategy exploits directly the very nature of transformers (separated task information from the pixels information) to tackle catastrophic forgetting with three components: (1) a task token expansion, (2) a divergence classifier, and (3) independent classifiers. All three greatly improve over the baseline transformer ($42.21\% \rightarrow 52.34\%$ in ``Last'') while having almost no memory overhead ($+0.2\%$).
The divergence classifier improves the diversity between task tokens: we observed that the minimal Euclidean distance between them increases by 8\%. Moreover, we also remarked that having independent classifiers reduces the Chaudhry et al.'s forgetting \cite{chaudhry2018riemannien_walk} by more than 24\%.

\input{tables/ablations}

%% file: tables/transf_archi.tex
\begin{table}[t]
\centering
\begin{tabular}{@{}l|cc@{}}
\hline
Hyperparameter & CIFAR & ImageNet \Tstrut\Bstrut\\
\hline
\# SAB & \multicolumn{2}{c}{5}\\
\# CAB & \multicolumn{2}{c}{1}\\
\# Attentions Heads & \multicolumn{2}{c}{12}\\
Embed Dim & \multicolumn{2}{c}{384}\\
Input Size & 32 & 224\\
Patch Size & 4 & 16 \\
\hline
\end{tabular}
\caption{\textbf{DyTox's architectures} for CIFAR and ImageNet. The only difference between the two architectures is the patch size, as the image sizes vary between datasets.}
\label{tab:archi}
\end{table}

%% file: tables/imagenet.tex
\begin{table*}[t]
	\centering
			\begin{tabular}{l|ccccc|ccccc}
			\toprule[0.3mm]
			 & \multicolumn{5}{c|}{ImageNet100 10 steps} & \multicolumn{5}{c}{ImageNet1000 10 steps} \\
			\cmidrule{2-11}
			 & \multirow{2}{*}{\textbf{$\#$P}} & \multicolumn{2}{c}{\textbf{top-1}}  & \multicolumn{2}{c|}{\textbf{top-5}} & \multirow{2}{*}{\textbf{$\#$P}}   & \multicolumn{2}{c}{\textbf{top-1}} & \multicolumn{2}{c}{\textbf{top-5}} \\
			\cmidrule{3-6}
			\cmidrule{8-11}
			\textbf{Methods} &   & \textbf{Avg}           & \textbf{Last}        & \textbf{Avg}     & \textbf{Last}   &     & \textbf{Avg}    & \textbf{Last}      & \textbf{Avg}     & \textbf{Last}      \\
			\hline
			ResNet18 joint  & $11.22$  & - & -  & - & $95.10$ & $11.68$ & - & - & - & $89.27$ \\
			Transf. joint & 11.00 & - & 79.12 & - & 93.48 & 11.35 & - & 73.58 & - & 90.60\\
			\midrule
			\textit{E2E} \cite{castro2018end_to_end_inc_learn} & 11.22 & - & - & 89.92 & 80.29 & 11.68 & - & - & 72.09 & 52.29 \\
			\textit{Simple-DER} \cite{li2021preserve} & - & - & - & - & - & 28.00 & 66.63 & 59.24 & 85.62 & 80.76 \\
			iCaRL \cite{rebuffi2017icarl} & $11.22$ & - & - & $83.60$ & $63.80$ & $11.68$ & $38.40$ & $22.70$& $63.70$ & $44.00$ \\
			BiC \cite{hou2019ucir} & $11.22$ & - & - & $90.60$ & $84.40$ & $11.68$ & - & - & $84.00$ & $73.20$\\
			WA \cite{zhao2020weightalignement} & $11.22$ & - & - & $91.00$ & $84.10$ & $11.68$ & $65.67$ & $55.60$ & $86.60$ & $81.10$ \\
			RPSNet \cite{rajasegaran2019rpsnet} & & - & - & $87.90$ & $74.00$ & - & - & - & - & - \\
			DER w/o P \cite{yan2021der}  & 112.27 & \textbf{77.18} & 66.70  & \textbf{93.23} & 87.52  & 116.89 & 68.84 & 60.16 & 88.17 & 82.86  \\ 
			\textcolor{gray}{$\text{DER}^\dagger$} \cite{yan2021der} & \textcolor{gray}{-} & \textcolor{gray}{76.12}  & \textcolor{gray}{66.06} & \textcolor{gray}{92.79} & \textcolor{gray}{88.38}  & \textcolor{gray}{-} & \textcolor{gray}{66.73} & \textcolor{gray}{58.62}  & \textcolor{gray}{87.08} & \textcolor{gray}{81.89}\\
			\hline
			DyTox & 11.01 & 77.15 & 69.10 & 92.04 & 87.98 & 11.36 & 71.29 & 63.34 & 88.59                           & 84.49\\
			\hline
			\end{tabular}
	\caption{\textbf{Results on ImageNet-100  and ImageNet-1000 datasets}, learned with 10 steps of respectively 10 and 100 new classes. E2E \cite{castro2018end_to_end_inc_learn} and Simple-DER \cite{li2021preserve} results come from their respective papers, and used a different class ordering. Other results come from \cite{yan2021der}. The $\dagger$ symbol means that \cite{yan2021der} needed setting-sensitive hyperparameters. Moreover, its reported parameters count was an average over all steps (\cite{yan2021der} reported 14.52M on ImageNet1000): the final parameters count (necessarily higher) was not available.}
	\label{tab:imagenet}
\end{table*}

%% file: tables/cifar100_b0.tex
\begin{table*}[t]
\centering
\resizebox{1.0\textwidth}{!}{%
\begin{tabular}{@{}l|ccc|ccc|ccc}
\hline
 & \multicolumn{3}{c}{10 steps} & \multicolumn{3}{c}{20 steps} & \multicolumn{3}{c}{50 steps}\\
\textbf{Methods}  & \textbf{\#P} & \textbf{Avg} & \textbf{Last} & \textbf{\#P} & \textbf{Avg} & \textbf{Last} & \textbf{\#P} & \textbf{Avg} & \textbf{Last}\\
\hline
ResNet18 Joint &  11.22 & - & 80.41 & 11.22 & - & 81.49 & 11.22 & - & 81.74\\
Transf. Joint & 10.72 & - & 76.12 & 10.72 & - & 76.12 & 10.72 & - & 76.12\\
\hline
iCaRL \cite{rebuffi2017icarl} & 11.22 & 65.27\scriptsize{\mypm1.02} & 50.74  & 11.22 & 61.20\scriptsize{\mypm0.83} & 43.75 & 11.22 & 56.08\scriptsize{\mypm0.83} & 36.62\\
UCIR \cite{hou2019ucir} & 11.22 & 58.66\scriptsize{\mypm0.71} & 43.39 & 11.22 & 58.17\scriptsize{\mypm0.30} & 40.63 & 11.22 & 56.86\scriptsize{\mypm0.83} & 37.09\\
BiC \cite{wu2019bias_correction} & 11.22 & 68.80\scriptsize{\mypm1.20} & 53.54 & 11.22 & 66.48\scriptsize{\mypm0.32} & 47.02 & 11.22 & 62.09\scriptsize{\mypm0.85} & 41.04\\
WA \cite{zhao2020weightalignement}  & 11.22 & 69.46\scriptsize{\mypm0.29} & 53.78 & 11.22 & 67.33\scriptsize{\mypm0.15} & 47.31 & 11.22 & 64.32\scriptsize{\mypm0.28} & 42.14\\
PODNet \cite{douillard2020podnet}  & 11.22 & 58.03\scriptsize{\mypm1.27} & 41.05 & 11.22 & 53.97\scriptsize{\mypm0.85} & 35.02 & 11.22 & 51.19\scriptsize{\mypm1.02} & 32.99\\
RPSNet \cite{rajasegaran2019rpsnet}  & 56.5\,\, & 68.60 & 57.05 & - & - & - & - & - & - \\
DER \small{w/o P} \cite{yan2021der}  & 112.27 & 75.36\scriptsize{\mypm0.36} & 65.22 & 224.55 & 74.09\scriptsize{\mypm0.33} & 62.48 & 561.39 & 72.41\scriptsize{\mypm0.36} & 59.08\\ 
\textcolor{gray}{$\text{DER}^\dagger$} \cite{yan2021der}  & \textcolor{gray}{-} & \textcolor{gray}{74.64\scriptsize{\mypm0.28}} & \textcolor{gray}{64.35} & \textcolor{gray}{-} & \textcolor{gray}{73.98\scriptsize{\mypm0.36}} & \textcolor{gray}{62.55} & \textcolor{gray}{-} & \textcolor{gray}{72.05\scriptsize{\mypm0.55}} & \textcolor{gray}{59.76}\\
\hline
DyTox & 10.73 & 73.66\mysmpm{0.02} & 60.67\mysmpm{0.34} & 10.74 & 72.27\mysmpm{0.18} & 56.32\mysmpm{{0.61}} & 10.77 & 70.20\mysmpm{0.16} & 52.34\mysmpm{0.26}\\
DyTox+ & 10.73 & 75.54\mysmpm{0.10} & 62.06\mysmpm{0.25} & 10.74 & 75.04\mysmpm{0.11} & 60.03\mysmpm{0.45} & 10.77 & 74.35\mysmpm{0.05} & 57.09\mysmpm{0.13}\\
\hline
\end{tabular}
}
\caption{\textbf{Results on CIFAR100} averaged over three different class orders. Baselines results are come from \cite{yan2021der}. The $\dagger$ symbol means that \cite{yan2021der} needed setting-sensitive hyperparameters. Moreover, its reported parameters count was an average over all steps: the final parameters count (necessarily higher) was not available.}
\label{tab:cifar100-b0}
\end{table*}

%% file: tables/plus.tex
\begin{table}[t]
\centering
\begin{tabular}{@{}l|c|cc}
\hline
 & 1 step & \multicolumn{2}{c}{50 steps}\\
\textbf{Training} & \textbf{Last} ($\uparrow$) & \textbf{Last} ($\uparrow$) & \textbf{Forgetting} ($\downarrow$)\Tstrut\Bstrut\\
\hline
DyTox & 76.12 & 52.34 & 33.15 \Tstrut\\
DyTox+ & 77.51\scriptsize{\textcolor{OliveGreen}{+1.39}} &  57.09\scriptsize{\textcolor{OliveGreen}{+4.75}} & 31.50\scriptsize{\textcolor{OliveGreen}{-1.65}}\Bstrut\\
\hline
\end{tabular}
\caption{\textbf{``Last'' accuracy and forgetting} \cite{chaudhry2018riemannien_walk} on CIFAR100 for the joint (1 step, no continual) and 50 steps settings.\vspace{-1em}}
\label{tab:training_plus}
\end{table}

%% file: tables/ablations.tex
\begin{table}
\centering
\begin{tabular}{ll|ccccc|cc}
& & \rot{\footnotesize{Knowledge Distillation}} & \rot{\footnotesize{Finetuning}} & \rot{\footnotesize{Token Expansion}} & \rot{\footnotesize{Divergence Classifier}} & \rot{\footnotesize{Indendepent Classifiers}} &  \textbf{Avg} & \textbf{Last}\\
\hline
\parbox[t]{2mm}{\multirow{7}{*}{\rotatebox[origin=c]{90}{\textbf{DyTox}}}} 
 & \parbox[t]{3mm}{\multirow{3}{*}{\rotatebox[origin=c]{90}{\textbf{\scriptsize{Transformer}}}}}
    &  &  &  &  &  & 60.69 & 38.87\Tstrut \\
   & &\cmark &  &  &  &  & 61.62 & 39.35 \\
& &\cmark & \cmark &  &  &  & 63.42 & 42.21 \\[3pt]
\cline{2-9} 
 & \parbox[t]{3mm}{\multirow{3}{*}{\rotatebox[origin=c]{90}{\textbf{\scriptsize{Dynamic}}}}}
    & \cmark & \cmark & \cmark &  &  & 67.30 & 47.57\Tstrut \\
    &&\cmark & \cmark & \cmark & \cmark &  & 68.28 & 49.45 \\
    &&\cmark & \cmark & \cmark & \cmark & \cmark & 70.20 & 52.34\Bstrut \\
\hline
\end{tabular}
\caption{\textbf{Ablations} of the different key components of our DyTox architecture. We report the average accuracy and the last accuracy on CIFAR100 for the setting with 50 steps.\vspace{-1em}}
\label{tab:ablation}
\end{table}

%% file: content/conclusion.tex

\section{Conclusion}

In this paper, we propose DyTox, a new dynamic strategy for continual learning based on transformer architecture. In our model, self-attention layers are shared across all tasks, and we add task-specific tokens to achieve task-specialized embeddings through a new task-attention layer. This architecture allows us to dynamically process new tasks with very little memory overhead and does not require complex hyperparameter tuning.
 Our experiments show that our framework scales to large datasets like ImageNet1k with state-of-the-art performances. Moreover, when a large number of tasks is considered (\textit{i.e.} CIFAR100 50 steps) our number of parameters increases reasonably contrary to previous dynamic strategies.

\vspace{0.1cm}\noindent \textbf{Limitations:} True continual learning aims at learning an almost unlimited number of tasks with low forgetting. No current approaches are yet able to do so. Thus, forgetting is not yet solved for continual learning but our model is a step forward in that direction.

\noindent \textbf{Broader impact:} Machine learning models often are biased, with some classes suffering from lower performances. Studying forgetting in continual learning provides insights about the difference in performances between classes. Our task-specialized model could help reduce these biases.

%% file: content/appendix.tex
\appendix
\section{Appendix}

\noindent\autoref{tab:notation} summarizes the notations used along this paper.

\input{tables/notation}

\subsection{Experimental details}

\paragraph{Datasets} We use three datasets: CIFAR100 \cite{krizhevskycifar100}, ImageNet100, and ImageNet1000 \cite{deng2009imagenet}. CIFAR100 is made of 50,000 train RGB images and 10,000 test RGB images of size $32\times32$ for 100 classes. ImageNet1000 contains 1.2 million RGB train images and 50,000 validation RGB images of size $224\times224$ for 1000 classes. ImageNet100 is a subset of 100 classes from ImageNet1000. We follow PODNet \cite{douillard2020podnet} and DER \cite{yan2021der} and use the same 100 classes they've used. Fine details about the datasets, like the class orders, can be found in the provided code in the options files (see readme).

\paragraph{Implementation} For all datasets, we train the model for 500 epochs per task with Adam \cite{kingma2014adam} with a learning rate of $5e^{-4}$, including 5 epochs of warmup.
Following UCIR \cite{hou2019ucir}, PODNet \cite{douillard2020podnet}, and DER \cite{yan2021der}, at the end of each task (except the first) we finetuned our model for 20 epochs with a learning rate of $5e^{-5}$ on a balanced dataset.
In DyTox, we applied the standard data augmentation of DeiT \cite{touvron2021deit} but we removed the pixel erasing \cite{zhong2017erasing}, MixUp \cite{hingyi2018mixup}, and CutMix \cite{yun2019cutmix} augmentations for fair comparison. In contrast, in DyTox+ we used a MixUp \cite{hingyi2018mixup} with beta distribution $\beta(0.8, 0.8)$. During all incremental tasks ($t>1$), the old classifiers $\operatorname{Clf}_i,\, i < t$ and the old task tokens $\theta_i,\, i < t$ parameters are frozen. During the finetuning phase where classes are rebalanced \cite{castro2018end_to_end_inc_learn,hou2019ucir,douillard2020podnet,yan2021der}, these parameters are optimized, but the SABs are frozen.

\paragraph{Hyperparameter tuning} In contrast with previous works \cite{douillard2020podnet,yan2021der}, we wanted stable hyperparameters, tuned for a single setting and then applied on all experiments. This avoids optimizing for the number of tasks, which defeats the purpose of continual learning \cite{farquhar2018robustcontinual}. We tuned hyperparameters for DyTox using a validation subset made of 10\% of the training dataset, and this only on the CIFAR100 experiment with 10 steps. We provide in \autoref{tab:tuning} the chosen hyperparameters. Results in the main paper shows that those hyperparameters reach state-of-the-art on all other settings and notably on ImageNet. 

\input{tables/tuning}

\paragraph{Baselines} E2E \cite{castro2018end_to_end_inc_learn} and Simple-DER \cite{li2021preserve} results come from their respective papers. All other baseline results are taken from the DER paper \cite{yan2021der}. We now further describe their contributions. iCaRL \cite{rebuffi2017icarl} uses a knowledge distillation loss \cite{hinton2015knowledge_distillation} and at test-time predicts using a k-NN from its features space. E2E \cite{castro2018end_to_end_inc_learn} learns a model with knowledge distillation and applies a finetuning after each step. UCIR \cite{hou2019ucir} uses cosine classifier and euclidean distance between the final flattened features as a distillation loss. BiC \cite{wu2019bias_correction} uses a knowledge distillation loss and also re-calibrates \cite{guo2017miscalibration} the logits of the new classes using a simple linear model trained on validation data. WA \cite{zhao2020weightalignement} uses a knowledge distillation loss and re-weights at each epoch the classifier weights associated to new classes so that they have the same average norm as the classifier weights of the old classes. PODNet \cite{douillard2020podnet} uses a cosine classifier and a specific distillation loss (POD) applied at multiple intermediary features of the ResNet backbone. RPSNet \cite{rajasegaran2019rpsnet} uses knowledge distillation and 
manipulates subnetworks in its architecture, following the lottery ticket hypothesis \cite{frankle2019lottery_ticket}. DER \cite{yan2021der} creates a new ResNet per task. All ResNets' embeddings are concatenated and fed to a unique classifier. ResNets are pruned using HAT \cite{serra2018hat} masking procedure. Note that DER pruning has multiple hyperparameters that are set differently according to the settings. Furthermore, the reported parameters count, after pruning, in \cite{yan2021der} is an average of the count over all steps: the final parameters count (necessarily higher) wasn't available. Finally, Simple-DER \cite{li2021preserve} is similar to DER, with a simpler pruning method which doesn't require any hyperparameter tuning.


\input{tables/alternatives}

\input{tables/cifar_plusplus}
\input{tables/imagenet100_plusplus}

\subsection{Parameter sharing of the TAB}

Previous dynamic methods as DER \cite{yan2021der} and Simple-DER \cite{li2021preserve} shared no parameters between tasks until the final classifier. We proposed instead with DyTox to share the encoder (SABs) and the decoder (TAB) parameters across tasks, leading to a minimal memory overhead while also sharing common information between tasks. In \autoref{tab:alternatives}, we compare the impact of sharing the TAB per task --- and only maintain different tokens per task. In the first row, a different TAB is created per task, while in the second row the same TAB is used --- which is the DyTox strategy.
A different TAB per task leads to better results (56\% \textit{v.s.} 52\% in ``Last'' accuracy) because the network can be more diverse with each TAB specialized to its associated task. This increased diversity has a drawback: the memory overhead is too important (97M \textit{v.s.} 10M parameters). We find in practice that DyTox strikes a good balance between memory overhead and continual performance.

\subsection{Novel continual training procedure}
\label{sec:plusplus}


\paragraph{DyTox++} We nicknamed DyTox+ our model when combined with a novel continual procedure based on MixUp \cite{hingyi2018mixup}. We now refine DyTox+ into DyTox++ by adding a new component during the training: the Sharpness-Aware Minimizer (SAM) \cite{kwon2021asam}. Indeed, \textbf{aiming for wider minima} is particularly important in continual learning \cite{kirkpatrick2017ewc,verwimp2021rehearsalrevealed}. This is because sharp task-specific minima lead to over-specialization to a particular task and consequently to a forgetting of all other tasks. Weights constraints as EWC \cite{kirkpatrick2017ewc} or second-order optimization \cite{lee2020kroneckercontinual} have similar motivations. SAM estimates the worst closest parameters during a first forward/backward pass, and then optimizes the loss w.r.t. to them during a second forward/pass. In consequence, DyTox++ optimizes the loss not w.r.t. the current parameters but w.r.t. a region of possible parameters leading to wide local minima that span across multiple tasks. In practice, we used the Adaptive SAM (ASAM) \cite{kwon2021asam}, an extension of SAM that is more robust to hyperparameters.

\paragraph{DyTox+ and DyTox++ experiments} The computational overhead of ASAM is lower than more complex second-order methods, but it still doubles the number of forward and backward passes. For this reason, we didn't include it in our main experiment but propose in \autoref{tab:cifar100-b0_pp} and \autoref{tab:imagenet_pp} experiments on CIFAR100 and ImageNet100. The gain provided by MixUp then ASAM on our model (DyTox++) leads to a consistent improvement of +4.7\% in ``Avg`` compared to the previous State-of-the-Art DER \cite{yan2021der} on CIFAR100 50 steps (\autoref{tab:cifar100-b0_pp} and +4.6\% on ImageNet100 10 steps (\autoref{tab:imagenet_pp}). Future works could consider the promising Look-SAM \cite{liu2021looksam} to reduce the time overhead.

\paragraph{Training procedure introspection} In \autoref{tab:training_plus}, we compare DyTox+ and DyTox++ on CIFAR100 in a joint setting (no continual) and in a continual setting with 50 steps. In the joint setting, our model slightly benefits from both MixUp and ASAM: the gain is limited (+1.79 \textit{p.p.}). On the other hand, those two methods greatly improve the extreme continual setting of 50 steps (+6.42 \textit{p.p.}). This shows that the gain is not due to absolute improvements of the model performance. Moreover, using the Chaudrhy et al.'s forgetting \cite{chaudhry2018riemannien_walk} measure, we compare how much a model has forgotten relatively to its previous tasks. This metric is therefore agnostic to absolute performance improvements. DyTox had a forgetting of 33.15\%, DyTox+ of 31.50\%, and DyTox++ of 30.47\%: a total reduction of 2.68 \pp. This validates our novel training procedures that are particularly efficient for continual learning.

\input{tables/plusplus}

\subsection{Patch size effect on forgetting}

Our model is the first application of transformers for continual computer vision. A key component of the transformer architecture is the patch tokenizer. The number of patch tokens in an image is determined by the patch size: a larger patch size means less tokens, and vice-versa. We wondered about the effect of the patch size on forgetting and tested three different kind of patch sizes in \autoref{tab:patch_size}. Echoing results in vision transformers \cite{dosovitskiy2020vit,touvron2021deit}, a smaller patch size (4 vs. 8 and 16) performs best in a joint training. However, the forgetting defined by Chaudhry et al. \cite{chaudhry2018riemannien_walk} is relatively similar, with 33.15\% for a patch of size of 4, and 33.20\% for a patch size of 16. Therefore, we argue that the transformer architecture is hardly sensitive to the patch resolution w.r.t its forgetting in continual learning.

\input{tables/patch_size}

\subsection{ResNet backbone}

DyTox is made of two main components: the SABs and the unique TAB. The TAB structure, taking in input both patch tokens and a task token, is crucial to our strategy. Yet, the SAB could be of any kind of features extractor, based on convolutions or attentions. Following the hybrid network proposed in ablations by Dosovitskiy et al. \cite{dosovitskiy2020vit}, we tried to replace the collection of SABs by a ResNet18. The final features of the ResNet, before global pooling, of shape $(W \times H \times D)$ can be seen as $W \times H$ tokens of $D$ dimension. We made a few modifications to this ResNet to boost its performance, namely removed the fourth and ultimate layer, and added a pointwise convolution with 504 output channels (so it can be divided by the 12 attention heads of the TAB), a batch normalization \cite{ioffe2015batchnorm}, and a ReLU activation.
These simple modifications are sufficient for our proof of concept, and thus we also didn't tune deeply this model. We display in \autoref{tab:resnet} the comparison of the two backbones on CIFAR100 50 steps: (1) with ResNet, and (2) with SABs (DyTox). Performances are slightly lower than DyTox with SABs, however, they are still significantly higher than previous state-of-the-art like WA \cite{zhao2020weightalignement}, especially in ``Last'' accuracy. Moreover, the parameters count is comparable to DyTox. This experiment shows that our DyTox framework, while designed with a transformer backbone in mind, is also efficient on non-token-based architectures such as a ResNet.

\input{tables/resnet_backbone}

\subsection{Alternative task decoders}

We investigate here other approaches for \textit{conditioning} features to the different tasks. Residual Adapters \cite{rebuffi2017residualadapters} adds a different residual branch made of a pointwise convolution for each domain the model is learned (\eg CIFAR then ImageNet then SVHN). This model needs the task/dataset/domain identifier at test-time to determine which residual to use. For VQA task \cite{antol2015vqa}, FiLM \cite{perez2018film} proposes to modify the visual features using the the textual query.
 
We adapt these two feature conditioning strategies for our transformer backbone architecture. We perform a global token pooling after the ultimate SAB, and apply for each learned task, a residual adapter or a FiLM. Residual adapter in our case is a MLP, and FiLM parameters are directly learned. As for DyTox, we forward each task-specific embedding to the respective task-specific classifier. We showcase the continual performance in \autoref{tab:task_cond} on CIFAR100 50 steps and ImageNet100 10 steps. On the former dataset, smaller and easier,  the residual adapters and FiLM have similar performance as our TAB approach. On the other hand, as soon as the task complexity increases with the more detailed ImageNet100 dataset, FiLM and Residual adapter based conditioning strategies forget significantly more than our complete DyTox framework: TAB outperform the Residual Adapters by +2.98 \pp in ``Last'' top-5 accuracy and FiLM by +6.58 \pp. 

\input{tables/task_cond}

\section{Erratum}

In Continual Learning, usually features are extracted once from images without data augmentations, and a procedure (random, closest \cite{castro2018end_to_end_inc_learn}, or even iCaRL iterative selection ``herding'' \cite{rebuffi2017icarl}) is applied to select rehearsal images. In this section, we call this family of selections \textit{Global Memory}.

In DyTox, when running our algorithm on multiple GPUs for data parallelism, we actually extracted features with data augmentations, and furthermore each GPU extracted features using their own slightly different data augmentations. As a result, each GPU could select different samples, and thus the effective memory size (across all processes) could be up to $N$ times bigger than asked, with $N$ the number of GPUs. We call that \textit{Distributed Memory}.

Therefore, DyTox's results in this paper, while interesting, are not exactly comparable to the compared baselines. This section acts as an erratum. That said, we believe that this Distributed Memory have merits.

\subsection{Interest of the Distributed Memory}

\begin{itemize}
    \item In the first place, it has similarity with the Federated Learning literature. In our case, each process/GPU/machine has access to its own memory. Which would make sense, as more machines would also mean a bigger disk storage.
    \item Because each GPU extracts with different data augmentations, the overlap of chosen among GPUs is very low: on CIFAR100 with 2 GPUs, the overlap is usually between 1\% and 10\%. This means that very representative/useful examples will be selected twice (and thus seen more often). But also that each GPU, because of the iterative nature of the rehearsal gathering of iCaRL, will select data belonging to different modes, leading to increased diversity.
\end{itemize}

It results in a direct gain of performance on CIFAR as highlighted in \autoref{tab:cifar100-b0-distMem}.

\input{tables/distMem}

\subsection{Updated results}

Here are the updated results. Global Memory uses 20 samples per class in total (as other baselines), Distributed Memory also uses 20 samples per class, but split across GPUs. e.g. on CIFAR100 with 2 GPUs, each GPU only samples 10 samples per class at most. So the overall effective memory size constraint is respected.

Overall, DyTox keeps strong performance, more often than not above WA \cite{zhao2020weightalignement} the third contender. Moreover, DyTox+, our improved version also presented in the original paper version still reaches excellent performance, due to the forgetting reduction we explained before.

Gray color symbolizes results presented in the original paper version.

CIFAR experiments were run on 2 GPUs, all are with distributed memory and are shown in \autoref{tab:cifar100-b0-erratum}.

ImageNet experiments, shown in \autoref{tab:imagenet-erratum} in global memory were run with 8 GPUs, and with distributed memory with 4 GPUs (thus only 5 images/class/GPU which probably explains here the slightly lower results compared to global memory).

\input{tables/cifar_erratum}

\input{tables/imagenet_erratum}

%% file: tables/notation.tex
\begin{table}[t]
\centering
\begin{tabular}{@{}l|l@{}}
\hline
Symbol & Meaning\Tstrut\Bstrut \\
\hline
$(x_i^t, y_i^t)$ & Input sample \& its label from the $t^{th}$ task\Tstrut \\
$C^t$ & Label set of the $t^{th}$ task\\
$C^{1:t}$ & All labels from all seen tasks\\
$\theta_t$ & Task token of the $t^{th}$ task\\
$\operatorname{Clf}_t$ & Independent classifier of the $t^{th}$ task\\
$\operatorname{SAB}_l$ & $l^{th}$ Self-Attention Block \\
$\operatorname{TAB}$ & Task-Attention Block \\
\hline
\end{tabular}
\caption{\textbf{Notations} used in the paper.}
\label{tab:notation}
\end{table}

%% file: tables/tuning.tex
\begin{table}[t]
\centering
\resizebox{0.47\textwidth}{!}{%
\begin{tabular}{l|cc}
\hline
Hyperparameter & Range & Chosen value\Tstrut\Bstrut \\
\hline
Learning rate & $1e^{-3}$, $5e^{-4}$, $1e^{-4}$ & $5e^{-4}$\Tstrut\\
Epochs & 300, 500, 700 & 500 \\
$\lambda$ & 0.05, 0.1, 0.5 & 0.1 \\
CIFAR's patch size & 4, 8, 16 & 4\\
ImageNet's patch size & 14, 16 & 16\\
\hline
\end{tabular}
}
\caption{\textbf{Hyperparameters} that were tuned from the codebase of \cite{touvron2021deit}. We ran a gridsearch on CIFAR100 10 steps on a validation set made of 10\% of the training set, and kept fixed the chosen hyperparameters for all experiments (any number of steps and any datasets).}
\label{tab:tuning}
\end{table}

%% file: tables/alternatives.tex
\begin{table}[t]
\centering
\begin{tabular}{@{}c|ccc@{}}
\hline
\textbf{TAB parameter sharing?}  & \textbf{\#P} &  \textbf{Avg} & \textbf{Last}\Tstrut\Bstrut\\
\hline
\xmark & 97.59 & 72.20 & 56.00 \\

\cmark & 10.77 & 70.20 & 52.34 \\
\hline
\end{tabular}
\caption{\textbf{Investigation of the parameter sharing of TAB}. We report the ``Avg'' accuracy and the ``Last'' accuracy for the 50 steps setting on CIFAR100. The second row corresponds to DyTox.}
\label{tab:alternatives}
\end{table}

%% file: tables/cifar_plusplus.tex
\begin{table*}[t]
\centering
\resizebox{1.0\textwidth}{!}{%
\begin{tabular}{@{}l|ccc|ccc|ccc}
\hline
 & \multicolumn{3}{c|}{10 steps} & \multicolumn{3}{c|}{20 steps} & \multicolumn{3}{c}{50 steps}\\
\textbf{Methods}  & \textbf{\#P} & \textbf{Avg} & \textbf{Last} & \textbf{\#P} & \textbf{Avg} & \textbf{Last} & \textbf{\#P} & \textbf{Avg} & \textbf{Last}\\
\hline
ResNet18 Joint &  11.22 & - & 80.41 & 11.22 & - & 81.49 & 11.22 & - & 81.74\\
Transf. Joint & 10.72 & - & 76.12 & 10.72 & - & 76.12 & 10.72 & - & 76.12\\
\hline
WA \cite{zhao2020weightalignement}  & 11.22 & 69.46\scriptsize{\mypm0.29} & 53.78 & 11.22 & 67.33\scriptsize{\mypm0.15} & 47.31 & 11.22 & 64.32\scriptsize{\mypm0.28} & 42.14\\
DER \small{w/o P} \cite{yan2021der}  & 112.27 & 75.36\scriptsize{\mypm0.36} & \textbf{65.22} & 224.55 & 74.09\scriptsize{\mypm0.33} & \textbf{62.48} & 561.39 & 72.41\scriptsize{\mypm0.36} & \textbf{59.08}\\ 
\hline
DyTox & 10.73 & 73.66\mysmpm{0.02} & 60.67\mysmpm{0.34} & 10.74 & 72.27\mysmpm{0.18} & 56.32\mysmpm{{0.61}} & 10.77 & 70.20\mysmpm{0.16} & 52.34\mysmpm{0.26}\\
DyTox+ & 10.73 & 75.54\mysmpm{0.10} & 62.06\mysmpm{0.25} & 10.74 & 75.04\mysmpm{0.11} & 60.03\mysmpm{0.45} & 10.77 & 74.35\mysmpm{0.05} & 57.09\mysmpm{0.13}\\
DyTox++ & 10.73 & \textbf{77.10\mysmpm{0.08}} & 64.53\mysmpm{0.08} & 10.74 & \textbf{76.57\mysmpm{0.18}} & \textbf{62.44\mysmpm{0.22}} & 10.77 & \textbf{75.45\mysmpm{0.19}} & 58.76\mysmpm{0.28}\\
\hline
DyTox+ & 10.73 & \textbf{76.74}\mysmpm{1.08} & \textbf{67.04}\mysmpm{0.10} & 10.74 & \textbf{76.25}\mysmpm{0.30} & \textbf{62.85}\mysmpm{0.16} & 10.77 & \textbf{74.16}\mysmpm{1.89} & \textbf{59.10}\mysmpm{0.99}\\
DyTox++ & 10.73 & \textbf{77.01}\mysmpm{1.21} & \textbf{67.53}\mysmpm{0.37} & 10.74 & \textbf{76.81}\mysmpm{0.43} & \textbf{64.27}\mysmpm{0.81} & 10.77 & \textbf{75.53}\mysmpm{2.79} & \textbf{59.51}\mysmpm{1.61}\\
\hline
\end{tabular}}
\caption{\textbf{Results on CIFAR100} averaged over three different class orders. WA and DER w/o P results are reported from \cite{yan2021der}. DyTox+ uses MixUp in addition of the DyTox strategy, DyTox++ further adds a sharpness-aware minimization \cite{kwon2021asam}.}
\label{tab:cifar100-b0_pp}
\end{table*}

%% file: tables/imagenet100_plusplus.tex
\begin{table}[t]
	\centering
	\resizebox{0.47\textwidth}{!}{%
			\begin{tabular}{l|ccccc}
			\hline
			 \multirow{2}{*}{\textbf{Methods}} & \multirow{2}{*}{\textbf{$\#$P}} & \multicolumn{2}{c}{\textbf{top-1}}  & \multicolumn{2}{c}{\textbf{top-5}} \\
			 &   & \textbf{Avg}           & \textbf{Last}        & \textbf{Avg}     & \textbf{Last} \\
			\hline
			ResNet18 joint  & $11.22$  & - & -  & - & $95.1$\\
			Transf. joint & 11.00 & - & 79.12 & - & 93.48\\
			\hline
			WA \cite{zhao2020weightalignement} & $11.22$ & - & - & $91.00$ & $84.10$ \\
			DER w/o P \cite{yan2021der}  & 112.27 & 77.18 & 66.70  & 93.23 & 87.52\\
			\hline
			DyTox & 11.01 & 77.15 & 69.10 & 92.04 & 87.98 \\
			DyTox+ & 11.01 & 79.22 & 69.06 & 93.72 & 88.82 \\
			DyTox++ & 11.01 & \textbf{80.76} & \textbf{72.46} & \textbf{94.40} & \textbf{90.10} \\
			\hline
			\end{tabular}
	}
	\caption{\textbf{Results on ImageNet-100} with 10 steps of 10 new classes each. WA and DER w/o P results are reported from \cite{yan2021der}. DyTox+ uses MixUp in addition of the DyTox strategy, DyTox++ further adds a sharpness-aware minimizer.}
	\label{tab:imagenet_pp}
\end{table}

%% file: tables/plusplus.tex
\begin{table}[t]
\centering
\begin{tabular}{@{}l|c|ccc}
\hline
 & Joint (1 step) & \multicolumn{2}{c}{50 steps}\\
\textbf{Training} & \textbf{Last} ($\uparrow$) & \textbf{Last} ($\uparrow$) & \textbf{Forgetting} ($\downarrow$)\Tstrut\Bstrut\\
\hline
DyTox & 76.12 & 52.34 & 33.15 \Tstrut\\
DyTox+ & 78.86\scriptsize{\textcolor{OliveGreen}{+1.39}} &  59.10\scriptsize{\textcolor{OliveGreen}{+4.75}} & 24.81\scriptsize{\textcolor{OliveGreen}{-1.65}}\Bstrut\\
DyTox++ & 78.70\scriptsize{\textcolor{OliveGreen}{+0.40}} & 59.51\scriptsize{\textcolor{OliveGreen}{+1.67}} & 26.70\scriptsize{\textcolor{OliveGreen}{-1.03}} \\
\hline
\end{tabular}
\caption{\textbf{``Last'' accuracy and forgetting} \cite{chaudhry2018riemannien_walk} on CIFAR100 for the joint (1 step, no continual) and 50 steps settings.}
\label{tab:training_plus}
\end{table}

%% file: tables/patch_size.tex
\begin{table}[t]
\centering
\begin{tabular}{@{}c|c|cc@{}}
\hline
& Joint (1 steps) & \multicolumn{2}{c}{50 steps}  \\
\textbf{Patch size} & \textbf{Last} ($\uparrow$) & \textbf{Last} ($\uparrow$) & \textbf{Forgetting} ($\downarrow$)\Bstrut\\
\hline
4 & 76.12 & 52.34 & 33.15\Tstrut\\
8 & 67.65 & 43.93 & 35.44\\
16 & 50.15 & 31.49 & 33.20\\
\hline
\end{tabular}
\caption{\textbf{Patch size effect on continual} for the joint (1 step, no continual) and 50 steps settings on CIFAR100.
We choose a patch size of 4 for our main experiments: yet, it has only few impact on forgetting.}
\label{tab:patch_size}
\end{table}

%% file: tables/resnet_backbone.tex
\begin{table}[t]
\centering
\begin{tabular}{@{}l|cccc@{}}
\hline
Encoder & \textbf{\#P} & \textbf{Avg} &  \textbf{Last} \Tstrut\Bstrut\\
\hline
ResNet & 10.68 & 68.53 & 50.05\Tstrut\\
SABs & 10.77 & 70.20 & 52.34\Bstrut\\
\hline
\end{tabular}
\caption{\textbf{Hybrid network} on CIFAR100 50 steps. While the features extractor is made of SABs in DyTox, here we instead use a modified ResNet18. Our framework still works well with a convolution-based approach.}
\label{tab:resnet}
\end{table}

%% file: tables/task_cond.tex
\begin{table}[t]
\centering
\begin{tabular}{@{}l|cc|cc@{}}
\hline
&\multicolumn{2}{c}{CIFAR100} & \multicolumn{2}{c}{ImageNet100}
\\
&\multicolumn{2}{c}{Top-1} & \multicolumn{2}{c}{Top-5}
\\
Task decoder & Avg & Last & Avg & Last \\
\hline
Residual Adapters \cite{rebuffi2017residualadapters} & 70.00 & 52.38 & 91.25 & 85.00 \\
FiLM \cite{perez2018film} & 69.42 & 54.05 & 89.49 & 81.40 \\
TAB \textit{(ours)} & 70.20 & 52.34 & 92.04 & 87.98\\
\hline
\end{tabular}
\caption{\textbf{Alternative task conditioner} on CIFAR100 50 steps and ImageNet100 10 steps. While the simpler Residual Adapters and FiLM perform similarly to our TAB on CIFAR100, they forget significantly more on the complex ImageNet100.}
\label{tab:task_cond}
\end{table}

%% file: tables/distMem.tex
\begin{table*}[t]
\centering
\begin{tabular}{@{}lcc|cc|cc|cc}
\hline
 & & & \multicolumn{2}{c}{10 steps} & \multicolumn{2}{c}{20 steps} & \multicolumn{2}{c}{50 steps}\\
\textbf{Memory Type} & \textbf{\# Memory / GPU} & \textbf{Total effective \# memory}   & \textbf{Avg} & \textbf{Last}  & \textbf{Avg} & \textbf{Last} & \textbf{Avg} & \textbf{Last}\\
\hline
Global & 2,000 & 2,000 & 67.33 & 51.68 & 67.30 & 48.45  & 64.39 & 43.47\\
\hline
Distributed &  1,000 & 2,000 & \textbf{71.50} & \textbf{57.76} & \textbf{68.86} & \textbf{51.47} & \textbf{64.82} & \textbf{45.61}\\

\hline
\end{tabular}
\caption{\textbf{Results on CIFAR100 for Distributed vs Global Memory} when run with 2 GPUs. Global memory uses the same 2k samples across each GPU, while Distributed memory uses 1k samples per memory with potential overlap across sets.}
\label{tab:cifar100-b0-distMem}
\end{table*}

%% file: tables/cifar_erratum.tex
\begin{table*}[t]
\centering
\resizebox{1.0\textwidth}{!}{%
\begin{tabular}{@{}l|ccc|ccc|ccc}
\hline
 & \multicolumn{3}{c}{10 steps} & \multicolumn{3}{c}{20 steps} & \multicolumn{3}{c}{50 steps}\\
\textbf{Methods}  & \textbf{\#P} & \textbf{Avg} & \textbf{Last} & \textbf{\#P} & \textbf{Avg} & \textbf{Last} & \textbf{\#P} & \textbf{Avg} & \textbf{Last}\\
\hline
ResNet18 Joint &  11.22 & - & 80.41 & 11.22 & - & 81.49 & 11.22 & - & 81.74\\
Transf. Joint & 10.72 & - & 76.12 & 10.72 & - & 76.12 & 10.72 & - & 76.12\\
\hline
iCaRL \cite{rebuffi2017icarl} & 11.22 & 65.27\scriptsize{\mypm1.02} & 50.74  & 11.22 & 61.20\scriptsize{\mypm0.83} & 43.75 & 11.22 & 56.08\scriptsize{\mypm0.83} & 36.62\\
UCIR \cite{hou2019ucir} & 11.22 & 58.66\scriptsize{\mypm0.71} & 43.39 & 11.22 & 58.17\scriptsize{\mypm0.30} & 40.63 & 11.22 & 56.86\scriptsize{\mypm0.83} & 37.09\\
BiC \cite{wu2019bias_correction} & 11.22 & 68.80\scriptsize{\mypm1.20} & 53.54 & 11.22 & 66.48\scriptsize{\mypm0.32} & 47.02 & 11.22 & 62.09\scriptsize{\mypm0.85} & 41.04\\
WA \cite{zhao2020weightalignement}  & 11.22 & 69.46\scriptsize{\mypm0.29} & 53.78 & 11.22 & 67.33\scriptsize{\mypm0.15} & 47.31 & 11.22 & 64.32\scriptsize{\mypm0.28} & 42.14\\
PODNet \cite{douillard2020podnet}  & 11.22 & 58.03\scriptsize{\mypm1.27} & 41.05 & 11.22 & 53.97\scriptsize{\mypm0.85} & 35.02 & 11.22 & 51.19\scriptsize{\mypm1.02} & 32.99\\
RPSNet \cite{rajasegaran2019rpsnet}  & 56.5\,\, & 68.60 & 57.05 & - & - & - & - & - & - \\
DER \small{w/o P} \cite{yan2021der}  & 112.27 & 75.36\scriptsize{\mypm0.36} & 65.22 & 224.55 & 74.09\scriptsize{\mypm0.33} & 62.48 & 561.39 & 72.41\scriptsize{\mypm0.36} & 59.08\\ 
\textcolor{gray}{$\text{DER}^\dagger$} \cite{yan2021der}  & \textcolor{gray}{-} & \textcolor{gray}{74.64\scriptsize{\mypm0.28}} & \textcolor{gray}{64.35} & \textcolor{gray}{-} & \textcolor{gray}{73.98\scriptsize{\mypm0.36}} & \textcolor{gray}{62.55} & \textcolor{gray}{-} & \textcolor{gray}{72.05\scriptsize{\mypm0.55}} & \textcolor{gray}{59.76}\\
\hline
\rowcolor{Gray}
DyTox & 10.73 & 73.66\mysmpm{0.02} & 60.67\mysmpm{0.34} & 10.74 & 72.27\mysmpm{0.18} & 56.32\mysmpm{{0.61}} & 10.77 & 70.20\mysmpm{0.16} & 52.34\mysmpm{0.26}\\
\rowcolor{Gray}
DyTox+ & 10.73 & 75.54\mysmpm{0.10} & 62.06\mysmpm{0.25} & 10.74 & 75.04\mysmpm{0.11} & 60.03\mysmpm{0.45} & 10.77 & 74.35\mysmpm{0.05} & 57.09\mysmpm{0.13}\\
\hline
DyTox distMem & 10.73 & 71.50 & 57.76 & 10.74 & 68.86 & 51.47 & 10.77 & 64.82 & 45.61\\
DyTox+ distMem & 10.73 & 74.10 & 62.34 & 10.74 & 71.62 & 57.43 & 10.77 & 68.90 & 51.09\\
\hline
\end{tabular}
}
\caption{\textbf{Results on CIFAR100}. Updated version with an erratum for Global vs. Distributed memory of the table \autoref{tab:cifar100-b0}. Gray color symbolizes results presented in the original paper version where up to 2 times the amount of rehearsal samples was used.}
\label{tab:cifar100-b0-erratum}
\end{table*}

%% file: tables/imagenet_erratum.tex
\begin{table*}[t]
	\centering
			\begin{tabular}{l|ccccc|ccccc}
			\toprule[0.3mm]
			 & \multicolumn{5}{c|}{ImageNet100 10 steps} & \multicolumn{5}{c}{ImageNet1000 10 steps} \\
			\cmidrule{2-11}
			 & \multirow{2}{*}{\textbf{$\#$P}} & \multicolumn{2}{c}{\textbf{top-1}}  & \multicolumn{2}{c|}{\textbf{top-5}} & \multirow{2}{*}{\textbf{$\#$P}}   & \multicolumn{2}{c}{\textbf{top-1}} & \multicolumn{2}{c}{\textbf{top-5}} \\
			\cmidrule{3-6}
			\cmidrule{8-11}
			\textbf{Methods} &   & \textbf{Avg}           & \textbf{Last}        & \textbf{Avg}     & \textbf{Last}   &     & \textbf{Avg}    & \textbf{Last}      & \textbf{Avg}     & \textbf{Last}      \\
			\hline
			ResNet18 joint  & $11.22$  & - & -  & - & $95.10$ & $11.68$ & - & - & - & $89.27$ \\
			Transf. joint & 11.00 & - & 79.12 & - & 93.48 & 11.35 & - & 73.58 & - & 90.60\\
			\midrule
			\textit{E2E} \cite{castro2018end_to_end_inc_learn} & 11.22 & - & - & 89.92 & 80.29 & 11.68 & - & - & 72.09 & 52.29 \\
			\textit{Simple-DER} \cite{li2021preserve} & - & - & - & - & - & 28.00 & 66.63 & 59.24 & 85.62 & 80.76 \\
			iCaRL \cite{rebuffi2017icarl} & $11.22$ & - & - & $83.60$ & $63.80$ & $11.68$ & $38.40$ & $22.70$& $63.70$ & $44.00$ \\
			BiC \cite{hou2019ucir} & $11.22$ & - & - & $90.60$ & $84.40$ & $11.68$ & - & - & $84.00$ & $73.20$\\
			WA \cite{zhao2020weightalignement} & $11.22$ & - & - & $91.00$ & $84.10$ & $11.68$ & $65.67$ & $55.60$ & $86.60$ & $81.10$ \\
			RPSNet \cite{rajasegaran2019rpsnet} & & - & - & $87.90$ & $74.00$ & - & - & - & - & - \\
			DER w/o P \cite{yan2021der}  & 112.27 & \textbf{77.18} & 66.70  & \textbf{93.23} & 87.52  & 116.89 & 68.84 & 60.16 & 88.17 & 82.86  \\ 
			\textcolor{gray}{$\text{DER}^\dagger$} \cite{yan2021der} & \textcolor{gray}{-} & \textcolor{gray}{76.12}  & \textcolor{gray}{66.06} & \textcolor{gray}{92.79} & \textcolor{gray}{88.38}  & \textcolor{gray}{-} & \textcolor{gray}{66.73} & \textcolor{gray}{58.62}  & \textcolor{gray}{87.08} & \textcolor{gray}{81.89}\\
			\hline
			\rowcolor{Gray}
			DyTox & 11.01 & 77.15 & 69.10 & 92.04 & 87.98 & 11.36 & 71.29 & 63.34 & 88.59                           & 84.49\\
			\rowcolor{Gray}
			DyTox+ & 11.01 & 79.22 & 69.06 & 93.72 & 88.82 & 11.01 & --- & --- & --- & ---\\
			\hline
			DyTox globalMem & 11.01 & 71.85 & 57.94 & 90.72 & 83.52 & 11.36 & 68.14 & 59.75 & 87.03 & 82.93\\ 
			DyTox+ globalMem & 11.01 & \textbf{77.62} & 65.94 & 93.15 & 88.78 & 11.36 & \textbf{73.21} & \textbf{64.56} & \textbf{91.09} & \textbf{87.07}\\
			\hline
			DyTox distMem & 11.01 & 73.96 & 62.20 & 91.29 & 85.60 & 11.36 & --- & --- & --- & ---\\
			DyTox+ distMem & 11.01 & 77.15 & \textbf{67.70} & 93.17 & \textbf{89.42} & 11.36 & 70.88 & 60.00 & 90.53 & 85.25\\
			\hline
			\end{tabular}
	\caption{\textbf{Results on ImageNet-100  and ImageNet-1000 datasets}. Updated version with an erratum for Global vs. Distributed memory of the table \autoref{tab:imagenet}. Gray color symbolizes results presented in the original paper version where up to 4 times the amount of rehearsal samples was used.}
	\label{tab:imagenet-erratum}
\end{table*}